\documentclass{article}

\usepackage{iclr2027_conference,times}

\usepackage[utf8]{inputenc}
\usepackage[T1]{fontenc}
\usepackage{hyperref}
\usepackage{url}
\usepackage{booktabs}
\usepackage{amsfonts}
\usepackage{nicefrac}
\usepackage{microtype}
\usepackage{xcolor}
\usepackage{graphicx}
\usepackage{amsmath}
\usepackage{amssymb}
\usepackage{mathtools}
\usepackage{amsthm}
\usepackage{wrapfig}
\usepackage{verbatim}
 
\definecolor{citecolor}{HTML}{0071bc}
\hypersetup{
    colorlinks=true,
    linkcolor=red,
    anchorcolor=citecolor,
    citecolor=citecolor,
    pdftitle={CoRe-Gen: Robust Spectrum-to-Structure Generation under Imperfect Fingerprint Conditions},
    pdfauthor={Tianbo Liu, Chixiang Lu, Jing Hao, Hengyu Zhang, Lifei Wang, Haibo Jiang, Xiaojuan Qi}
}

\title{CoRe-Gen: Robust Spectrum-to-Structure Generation under Imperfect Fingerprint Conditions}

\author{
\textbf{Tianbo Liu}\textsuperscript{1}
\quad
\textbf{Chixiang Lu}\textsuperscript{1}
\quad
\textbf{Jing Hao}\textsuperscript{1}
\quad
\textbf{Hengyu Zhang}\textsuperscript{2}
\\
\textbf{Lifei Wang}\textsuperscript{3}
\quad
\textbf{Haibo Jiang}\textsuperscript{1}
\quad
\textbf{Xiaojuan Qi}\textsuperscript{1}
\\
\textsuperscript{1}The University of Hong Kong
\quad
\textsuperscript{2}The Chinese University of Hong Kong
\\
\textsuperscript{3}Zhejiang Shuren University
}

\iclrfinalcopy

\begin{document}

\maketitle

\begin{abstract}
Molecular structure elucidation from tandem mass spectra (MS/MS) remains challenging, particularly for de novo generation beyond database coverage. A common approach decomposes the task into spectrum-to-fingerprint prediction followed by fingerprint-to-structure decoding, enabling the use of large-scale molecular corpora. However, at deployment, the decoder relies on predicted rather than oracle fingerprints, introducing structured errors that propagate into generation. Moreover, common autoregressive decoders threshold fingerprint probabilities and serialize active bits as discrete input tokens, blocking molecular generation gradients from the spectrum encoder.

We present \textit{CoRe-Gen}, which improves the intermediate condition through synthetic-spectrum pretraining, matches deployment-time noise through frequency-aware corruption, and introduces a differentiable fingerprint-to-memory bridge for joint encoder--decoder finetuning. The bridge preserves soft bit confidences and conditions a structure-aware autoregressive decoder without discrete fingerprint tokenization. CoRe-Gen achieves 21.49\%/35.42\% Top-1/Top-10 exact-match accuracy on NPLIB1 and, under encoder-aligned evaluation, 20.02\%/22.45\% on MassSpecGym, while retaining efficient autoregressive inference.
\end{abstract}

\section{Introduction}
\label{intro}

Molecular structure elucidation from tandem mass spectrometry (MS/MS) is a fundamental problem in metabolomics, natural product discovery, and chemical analysis \citep{stein2012mass, horai2010massbank, guijas2018metlin}. Despite its importance, reliable spectrum-to-structure generation remains challenging due to the scarcity and bias of high-quality spectrum–molecule pairs, the variability of spectra across instruments and acquisition settings, and the strict chemical validity constraints imposed on generated molecules \citep{scheubert2013computational, stein2012mass, bushuiev2024massspecgym}. These challenges are further amplified in realistic settings where models must generalize beyond curated datasets to noisy, heterogeneous spectra encountered in practice.


Existing approaches broadly fall into two paradigms: retrieval-based identification and de novo generation. Retrieval methods \citep{duhrkop2015searching, wang2016sharing, spec2vec, wolf2010metfrag,
allen2014cfmid, wang2022cfmid4, brouard2016iokr, schymanski2017casmi2016,
horai2010massbank, guijas2018metlin} map spectra into a searchable space and rank candidates from reference databases, achieving strong performance when the target molecule or its analogues are covered. However, they are fundamentally limited by database completeness and are sensitive to experimental mismatch. In contrast, de novo generation methods \citep{litsa2023spec2mol, liu2023ms2smiles, stravs2022msnovelist, wang2025madgen,
butler2023ms2mol} aim to synthesize molecular structures directly from spectra, enabling discovery beyond existing databases. While recent neural models \citep{stravs2022msnovelist, litsa2023spec2mol, liu2023ms2smiles, han2025msbart, wang2025madgen, bohde2025diffms} have improved generation quality, they remain constrained by limited paired supervision and are often unstable under real-world spectral variability. 

\begin{figure}[!t]
  \centering
  \includegraphics[width=\columnwidth]{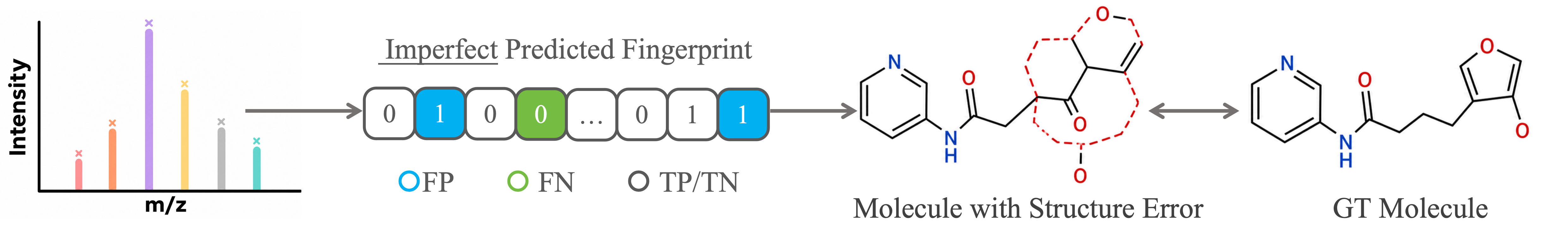}
  \caption{Imperfect Fingerprints Bottleneck Generation.}
  \label{fig:overview}
\end{figure}

A practical compromise is fingerprint-conditioned generation, where a spectrum encoder predicts a molecular fingerprint that conditions downstream structure decoding \citep{goldman2023annotating, ucak2020neuraldecipher, ucak2023reconstruction, stravs2022msnovelist,bohde2025diffms, han2025msbart}. While this decomposition enables large-scale fingerprint--structure pretraining, many autoregressive implementations, including MolForge-style decoders and MS-BART, threshold the predicted fingerprint and serialize active bit indices as input tokens \citep{ucak2023reconstruction,han2025msbart}. This hard interface discards bit confidence and prevents the molecular generation loss from propagating through the thresholding/index-selection step to the spectrum encoder. Meanwhile, scarce and biased spectrum--molecule supervision makes fingerprint predictions noisy, especially for long-tail bits, producing structured errors that propagate into generation (Figure~\ref{fig:overview}).

This exposes two coupled gaps: decoders are trained on clean fingerprints but deployed on noisy predictions, and their discrete condition interface prevents end-to-end correction of the upstream encoder. We therefore improve the predicted condition, match its corruption during training, preserve its soft confidence through a differentiable bridge, and exploit residual uncertainty during decoding.

To this end, we propose \textit{CoRe-Gen} to explicitly address imperfect fingerprint conditions in spectrum-to-structure generation. Our framework is built on four complementary principles.
(1) \textit{Improve condition quality}: we pretrain the spectrum encoder on large-scale synthetic spectra from approximately 790K valid organic molecules, each paired with five types of in-silico MS/MS spectra, to enhance fingerprint prediction under limited real supervision. While synthetic data is less suitable for end-to-end de novo generation due to the domain gap between simulated and real spectra, fingerprint prediction provides a coarse and redundant intermediate target that is substantially more robust to such imperfections, making it a more effective and transferable pretraining objective.
(2) \textit{Match the condition to reduce the train–test gap}: we introduce frequency-aware fingerprint corruption during decoder pretraining to simulate realistic prediction noise, thereby aligning training conditions with deployment-time inputs.
(3) \textit{Enable joint finetuning}: a differentiable fingerprint-to-memory bridge converts soft bit probabilities directly into continuous decoder memory, allowing sequence loss to update the spectrum encoder.
(4) \textit{Mitigate imperfect conditions via structural regularization}: we design a structure-aware autoregressive decoder that incorporates compositional SELFIES representations \citep{krenn2020selfies}, auxiliary structural supervision, and lightweight chemical constraints to stabilize generation under uncertainty.
By jointly addressing condition quality, condition mismatch, and robustness, CoRe-Gen transforms fingerprint-conditioned generation from an idealized assumption into a more practical and reliable pipeline. 

Extensive experiments demonstrate the effectiveness of CoRe-Gen. On NPLIB1, it reaches 21.49\%/35.42\% Top-1/Top-10 exact-match accuracy and 0.57/0.67 Tanimoto similarity. On MassSpecGym, an encoder-aligned comparison yields 20.02\%/22.45\% exact-match accuracy and 0.40/0.47 Tanimoto similarity. CoRe-Gen also preserves efficient autoregressive decoding, requiring approximately 2 seconds per spectrum for 100 candidates on a single RTX 4090 GPU, compared with approximately 160 seconds for DiffMS.

Our contributions are summarized as follows:

\begin{itemize}
  \item We identify imperfect predicted fingerprints as the central bottleneck in practical spectrum-to-structure generation and reformulate the problem as robust generation under condition mismatch.  
  \item  We propose a unified solution that improves fingerprint prediction via synthetic pretraining, matches deployment noise through corruption-aware training, and exploits imperfect conditions with structure-aware decoding. 
  \item We replace hard fingerprint tokenization with a differentiable fingerprint-to-memory bridge, enabling joint encoder--decoder finetuning while retaining autoregressive efficiency.
  \item CoRe-Gen reaches 21.49\%/35.42\% exact-match accuracy on NPLIB1 and 20.02\%/22.45\% on MassSpecGym at Top-1/Top-10.
\end{itemize}

\section{Related Work}

\noindent\textbf{Spectrum-to-Structure Elucidation.}
Spectrum-to-structure methods broadly follow retrieval-based identification or de novo
generation. Retrieval systems search spectral libraries or rank database candidates
using fragmentation and learned scoring, including GNPS, MassBank, METLIN,
SIRIUS/CSI:FingerID, MetFrag, CFM-ID, IOKR, and molDiscovery
\citep{stein2012mass, wang2016sharing, horai2010massbank, guijas2018metlin,
duhrkop2015searching, duhrkop2019sirius, wolf2010metfrag, allen2014cfmid,
wang2022cfmid4, brouard2016iokr, cao2021moldiscovery}. They are effective when relevant
candidates are covered, but remain limited by database coverage and experimental
mismatch \citep{schymanski2013casmi, schymanski2017casmi2016}. De novo methods instead
generate beyond observed libraries using sequence, retrieval-guided, or graph-based
models \citep{litsa2023spec2mol, liu2023ms2smiles, butler2023ms2mol,
wang2025madgen, bohde2025diffms, han2025msbart}, but remain sensitive to scarce paired
supervision and spectral variability.

Learned spectral embeddings and molecular fingerprints bridge these paradigms by
providing chemically meaningful intermediate targets for similarity learning and
downstream prediction \citep{spec2vec, huber2021ms2deepscore, li2021spectralentropy,
dejonge2023ms2query, goldman2023annotating, zhang2024msbert, wei2019neims,
ji2020deepei, young2024massformer}. We focus on fingerprint prediction not as an endpoint,
but as an imperfect intermediate condition for molecular generation.

\noindent\textbf{Fingerprint-Conditioned Generation under Imperfect Conditions.}
Fingerprint-conditioned generation is closest to our setting because it decouples
spectrum interpretation from molecular decoding. Neural Decipher and MSNovelist invert
ECFP-like fingerprints, while subsequent work combines fingerprint conditions with
pretraining, retrieval guidance, and graph-based generation
\citep{ucak2020neuraldecipher, stravs2022msnovelist, wang2025madgen,
bohde2025diffms, han2025msbart}. MolForge represents a binary fingerprint as the sequence
of active bit indices, and MS-BART similarly thresholds predicted probabilities before
mapping active bits to fingerprint tokens \citep{ucak2023reconstruction,han2025msbart}.
Although effective for autoregressive generation, this serialization discards confidence
and blocks gradients through the discrete interface. Our bridge instead maps all soft bit
probabilities to continuous decoder memory, enabling joint finetuning. We also address the
clean-to-predicted mismatch caused by structured false-positive and false-negative errors.
Graph generators can impose strong structural constraints but incur higher sampling
cost \citep{bohde2025diffms}. Autoregressive sequence models remain attractive at scale
because of their efficiency, but are more vulnerable to local error accumulation under
imperfect conditions. Our approach therefore adds lightweight chemical priors to
efficient autoregressive decoding while explicitly training for this mismatch.

\section{Method}
\label{sec:method}

Given an MS/MS spectrum $\mathbf{x}$, our goal is to generate the target molecular
sequence $\mathbf{y}$ through an intermediate molecular fingerprint
$\mathbf{f}\in\{0,1\}^d$. Fingerprint-conditioned generation decomposes the problem as
\begin{equation}
p_{\theta}(\mathbf{y}\mid \mathbf{x})
\approx
\sum_{\mathbf{f}} p_{\theta_e}(\mathbf{f}\mid \mathbf{x})\; p_{\theta_d}(\mathbf{y}\mid \mathbf{f}),
\label{eq:problem_factorization}
\end{equation}
where $\theta_e$ and $\theta_d$ denote encoder and decoder parameters, respectively.
At inference time, the decoder receives the predicted soft fingerprint
$\hat{\mathbf{f}}\in[0,1]^d$ rather than the oracle fingerprint. Instead of thresholding
it into a discrete index sequence, CoRe-Gen constructs continuous memory
$\mathbf{M}=h_{\phi}(\hat{\mathbf{f}})$ and optimizes generation through this bridge:
\begin{equation}
\max_{\theta_e,\phi,\theta_d}\;
\mathbb{E}_{(\mathbf{x},\mathbf{y})}
\left[\log p_{\theta_d}\!\left(\mathbf{y}\mid h_{\phi}(\hat{\mathbf{f}})\right)\right],
\qquad
\hat{\mathbf{f}} = g_{\theta_e}(\mathbf{x}).
\end{equation}

\begin{figure}[!t]
  \centering
  \includegraphics[width=\columnwidth]{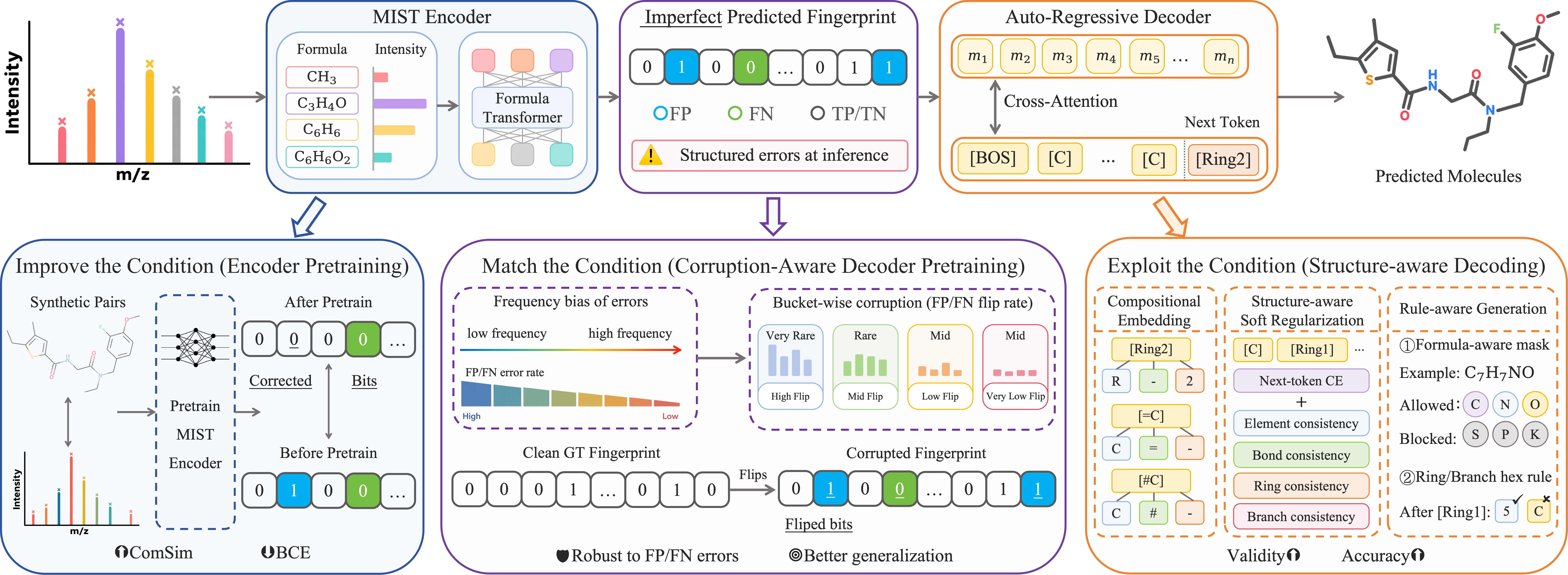}
  \caption{Overview of CoRe-Gen: improving, matching, and exploiting imperfect fingerprint conditions for robust spectrum-to-structure generation.}
  \label{fig:method_pipeline}
  \vspace{-0.9em}
\end{figure}

We propose CoRe-Gen, a condition-centric framework whose three robustness components are connected by a differentiable fingerprint interface. As illustrated in Figure~\ref{fig:method_pipeline}, we first \emph{improve the condition} through synthetic-spectrum encoder pretraining, then \emph{match the condition} by exposing the decoder to structured fingerprint errors, and finally \emph{exploit the imperfect condition} with structure-aware autoregressive decoding. The differentiable bridge below couples the encoder and decoder during final finetuning.

\textbf{Differentiable fingerprint-to-memory bridge.}
Prior sequence decoders commonly retain only thresholded active bits. We instead keep
all soft probabilities and fold the $d=4096$ dimensions into $C=64$ modulo groups. For
group $i$, learned bit embeddings are confidence-weighted and normalized,
$\mathbf{c}_i=\sum_k \hat f_{i+kC}\mathbf{e}_{i+kC}/(\sum_k\hat f_{i+kC}+\epsilon)$.
The 64 chunk features are contextualized into fixed-length memory tokens and consumed by
decoder cross-attention. Because the bridge contains no threshold or index-selection
operation, the autoregressive loss backpropagates through the memory tokens into the
spectrum encoder, enabling joint finetuning. Appendix~\ref{app:fp_memory_bridge} gives
the full architecture and illustration.

\subsection{Improve the Condition: Encoder Pretraining}
\begin{wrapfigure}[13]{r}{0.44\columnwidth}
  \vspace{-1.2em}
  \centering
  \includegraphics[width=0.42\columnwidth]{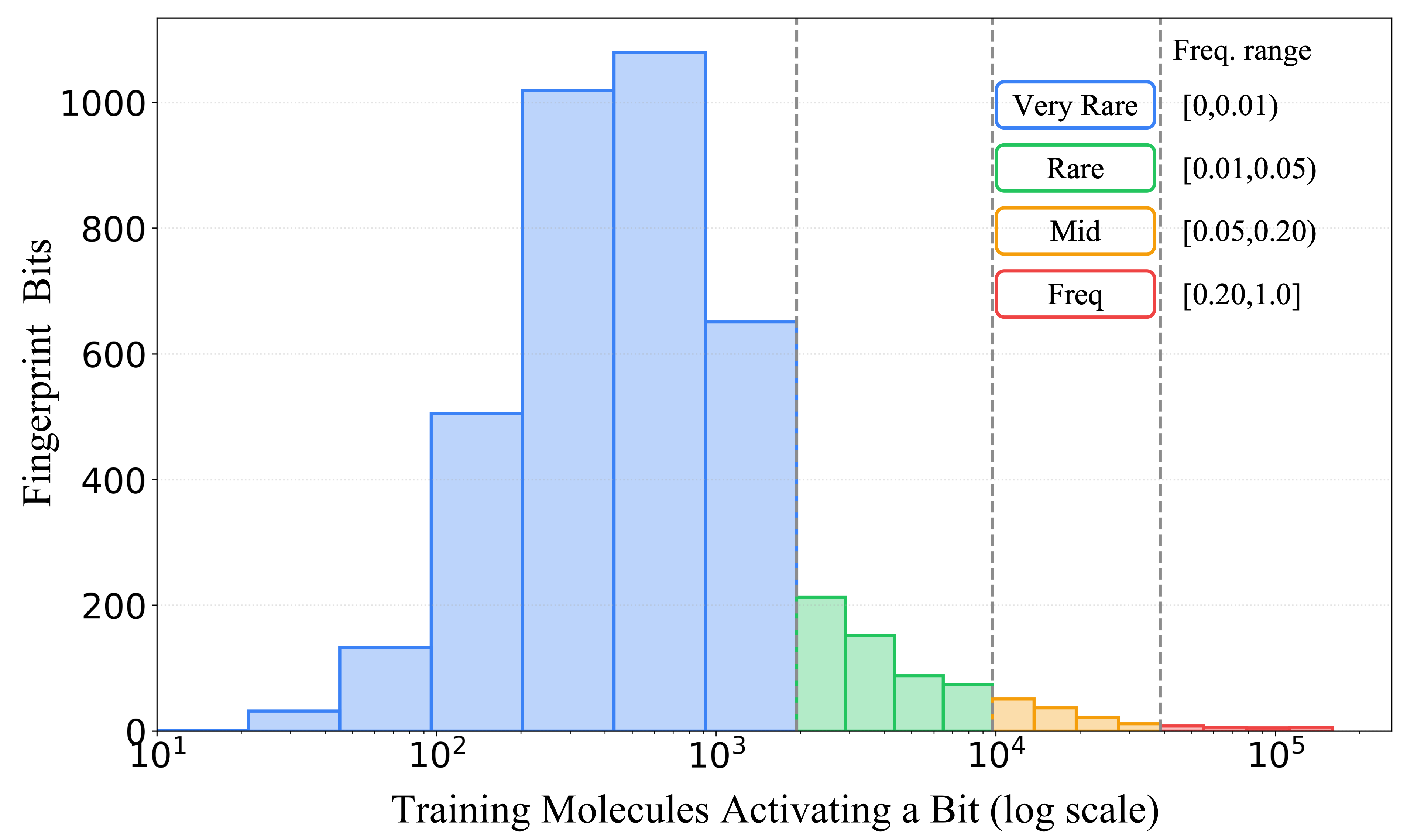}
  \caption{Long-tail distribution of active fingerprint bits in real paired data.}
  \label{fig:bit_freq}
  \vspace{-0.8em}
\end{wrapfigure}

We instantiate $p_{\theta_e}(\mathbf{f}\mid\mathbf{x})$ with the MIST spectrum encoder
and a lightly adapted FP-Growing head that predicts Morgan fingerprints with $d=4096$
bits. The main weakness of this spectrum-to-fingerprint interface is not architectural
capacity alone, but insufficient supervision for most fingerprint dimensions. As shown
in Figure~\ref{fig:bit_freq}, active bits in real paired datasets are highly
long-tailed: many substructure bits appear in only a tiny fraction of training spectra,
so the encoder receives few positive examples for learning them.

This long-tailed supervision directly degrades the predicted condition. Low-frequency
bits are systematically under-trained, making the encoder less reliable for rare but
chemically informative substructures. To alleviate this limitation, we introduce
large-scale synthetic-spectrum pretraining. The synthetic pretraining data come from a
large-scale simulated spectroscopic corpus containing approximately 790K valid organic
molecules~\cite{alberts2024unraveling}. Each molecule is associated with five types of
in-silico MS/MS spectra: CFM-ID 4.0 spectra at collision energies of 10, 20, and
40~eV~\cite{allen2014cfmid,duhrkop2019sirius}, together with spectra generated by
ICEBERG~\cite{goldman2023iceberg} and SCARF~\cite{goldman2023scarf}. The combination of
multiple simulators and collision-energy settings yields a substantially larger and more
diverse spectrum--molecule corpus than typical experimental MS/MS libraries.

Compared with the MassSpecGym training set, this corpus substantially densifies the
long-tail fingerprint space: fingerprint bits that occur at most 20 times in
MassSpecGym are observed on average 1,164.8 times in the pretraining set, and all
199 such rare bits receive at least a 10$\times$ increase in occurrence count. Although
simulated spectra are lower-fidelity than experimental spectra and may not fully capture
real instrument noise, peak intensities, or acquisition-specific artifacts, their
fragment peaks still provide useful chemical evidence for substructure-level fingerprint
bits. Therefore, synthetic data are well suited for pretraining spectrum encoders to
learn general spectrum-to-substructure associations, followed by fine-tuning on real
spectra for adaptation to experimental noise and acquisition conditions.

After pretraining, the encoder is fine-tuned on real spectra with the standard BCE
objective, with optional bit-frequency reweighting so that rare but informative bits are
not overwhelmed by frequent fingerprint dimensions. Table~\ref{tab:pretrain_effect}
reports the effect of encoder pretraining on both benchmarks. In our framework, encoder
pretraining matters because it improves the downstream usability of predicted
fingerprints, especially on weakly supervised long-tail bits.

\subsection{Match the Condition: Corruption-Aware Decoder Pretraining}
Decoder pretraining on clean fingerprint-molecule pairs creates a condition mismatch:
the decoder learns from oracle fingerprints but is deployed on predicted fingerprints.
We reduce this gap by corrupting clean fingerprints with structured errors that reflect
bit-frequency-dependent prediction behavior. Let $c_j$ denote the occurrence frequency
of bit $j$ in the training corpus. Motivated by the long-tailed distribution in
Figure~\ref{fig:bit_freq}, we partition the $d$ bits into four frequency buckets
$\{B_r\}_{r=1}^4$ by quantiles of $\{c_j\}_{j=1}^d$, corresponding to very-rare, rare,
mid, and frequent bits, and let $b(j)$ denote the bucket index of bit $j$.

\begin{wraptable}[8]{r}{0.4\columnwidth}   
  \vspace{-2em}
  \centering
  \caption{MassSpecGym bit-frequency buckets for encoder prediction.}
  \label{tab:bit_bucket_msg_compact}
  \scriptsize
  \setlength{\tabcolsep}{5pt}
  \begin{tabular}{lccc}
    \toprule
    Bucket & \# bits & Freq. range & Precision \\
    \midrule
    Very rare & 3422 & $[0,0.01)$ & 0.198 \\
    Rare & 527 & $[0.01,0.05)$ & 0.314 \\
    Mid & 122 & $[0.05,0.20)$ & 0.428 \\
    Frequent & 25 & $[0.20,1.0]$ & 0.692 \\
    \bottomrule
  \end{tabular}
  \vspace{-1em}
\end{wraptable}

Table~\ref{tab:bit_bucket_msg_compact} shows this structure on MassSpecGym:
very-rare bits dominate the fingerprint space but are predicted with much lower
precision than frequent bits. We therefore model corruption at the bucket level
instead of applying uniform random noise. For each bucket $B_r$, we estimate
false-negative and false-positive tendencies from training data only (never from a
downstream test set):
\begin{equation}
\eta_r^- = 1-\mathrm{Recall}(B_r),
\qquad
\eta_r^+ = 1-\mathrm{Precision}(B_r).
\end{equation}
To avoid unstable per-bit estimates and degenerate zero-probability buckets, we
convert these tendencies into bucket-level sampling weights:
\begin{equation}
w_r^\pm
=
\epsilon
+
\alpha
\cdot
\frac{
\eta_r^\pm - \min_s \eta_s^\pm
}{
\max_s \eta_s^\pm - \min_s \eta_s^\pm
},
\label{eq:bucket_weight}
\end{equation}
where $\epsilon$ is a small floor value and $\alpha$ controls the strength of
the frequency-aware corruption pattern.

For a training sample with active set
$A(\mathbf{f})=\{j:f_j=1\}$ and inactive set
$\bar{A}(\mathbf{f})=\{j:f_j=0\}$, we first draw a corruption gate
\begin{equation}
z \sim \mathrm{Bernoulli}(p_{\mathrm{corr}}).
\end{equation}
If $z=0$, we keep the fingerprint unchanged. Otherwise, we draw a corruption
budget
\begin{equation}
k \sim \mathrm{clip}\!\left(\mathrm{Poisson}(\lambda), k_{\min}, k_{\max}\right),
\qquad
k_{\mathrm{eff}}
=
\min\!\bigl(k,\; |A(\mathbf{f})|,\; |\bar{A}(\mathbf{f})|\bigr).
\label{eq:corruption_budget}
\end{equation}
In our implementation, $\lambda$ is set according to the maximum corruption
budget, i.e., $\lambda=k_{\max}/2$.

We then sample a drop set $S^-\subseteq A(\mathbf{f})$ and an add set
$S^+\subseteq \bar{A}(\mathbf{f})$, both of size $k_{\mathrm{eff}}$, using
the normalized bucket-level weights:
\begin{equation}
p_j^- =
\frac{w^-_{b(j)}}{\sum_{\ell\in A(\mathbf{f})} w^-_{b(\ell)}}\;(j\in A(\mathbf{f})),
\qquad
p_j^+ =
\frac{w^+_{b(j)}}{\sum_{\ell\in \bar{A}(\mathbf{f})} w^+_{b(\ell)}}\;(j\in \bar{A}(\mathbf{f})).
\label{eq:corruption_sampling}
\end{equation}
The corrupted fingerprint $\tilde{\mathbf{f}}$ is constructed by
\begin{equation}
\tilde{f}_j = \left\{0\ \text{if }j\in S^-;\quad
1\ \text{if }j\in S^+;\quad f_j\ \text{otherwise}\right\}.
\label{eq:bit_swap}
\end{equation}
This sparsity-preserving swap injects frequency-aware structured FP/FN errors. 
Then the decoder is pretrained to generate the target modified SELFIES sequence from
$\tilde{\mathbf{f}}$, so its conditioning distribution more closely matches
deployment-time predicted fingerprints.

\subsection{Exploit the Imperfect Condition: Structure-Aware AR Decoding}
Even after improving and matching the fingerprint condition, residual fingerprint errors
can still be amplified by token-level autoregressive decoding. The role of
structure-aware decoding is to stabilize local chemical recovery when the fingerprint
condition remains imperfect. We therefore add three lightweight structural priors to the
decoder: compositional SELFIES embeddings, auxiliary structural supervision, and
rule-aware logit masking.

\textbf{Compositional SELFIES embedding.}
Instead of treating SELFIES tokens as independent identities, we parse each token $t$
into a compositional tuple $(p_t,a_t,b_t,q_t,d_t)$ consisting of part type, element,
bond prefix, ring/branch order, and optional hex digit. This factorization injects
lightweight structural priors that are usually more explicit in graph-based models. For
example, tokens such as \texttt{C} and \texttt{=C} should share their carbon-atom
identity while preserving their different bond contexts. 
By sharing factor embeddings across such related tokens, the decoder can reuse
chemical regularities across the vocabulary while still distinguishing the factors that
matter for generation. This also reduces the effective token vocabulary and parameter
cost, improving computational efficiency. Appendix~\ref{app:modified_selfies} describes
the detailed modified SELFIES interface. The token embedding is then built additively:
\begin{equation}
\begin{aligned}
\mathbf{h}_t
=
&\;\mathbf{e}_{\text{part}}(p_t)
+ \mathbf{e}_{\text{elem}}(a_t)
+ \mathbf{e}_{\text{bond}}(b_t) \\
&\;+\; \mathbb{I}(q_t > 0)\, \mathbf{e}_{\text{rb}}(q_t)
+\; \mathbb{I}(d_t \ge 0)\, \mathbf{e}_{\text{digit}}(d_t) \\
&\;+\; \mathbb{I}\!\left(c(\tilde{y}_t)\ge\tau\right)\mathbf{e}_{\text{res}}(\tilde{y}_t),
\end{aligned}
\label{eq:structure_aware_embedding}
\end{equation}
where $\tilde{y}_t$ is a canonicalized token form (removing isotope/chirality/H/charge
markers while preserving bond prefix and element), $c(\cdot)$ is its corpus count, and
$\tau$ is a frequency threshold. The residual term is enabled only for high-frequency
canonical tokens, which preserves flexibility on common patterns without overfitting
long-tail symbols.

\textbf{Auxiliary structural supervision.}
The same molecule may admit multiple valid SELFIES strings, so pure token-level
cross-entropy can over-penalize sequence variants while missing higher-level structural
consistency. However, key molecular statistics are invariant across equivalent
representations: the molecular formula is fixed, and so are coarse structural counts
such as the number of branch controls, ring controls, and double or triple bonds. These
invariants give us an opportunity to supervise generation at the topological-structure
level rather than only at the next-token level. We therefore augment token
cross-entropy with auxiliary losses over the corresponding chemical factors:
\begin{equation}
\mathcal{L}_{\mathrm{dec}}
=
\mathcal{L}_{\mathrm{CE}}
\;+\;
\lambda_{\mathrm{sent}}
\left(
w_{\mathrm{elem}}\mathcal{L}_{\mathrm{elem}}
+
w_{\mathrm{bond}}\mathcal{L}_{\mathrm{bond}}
+
w_{\mathrm{ring}}\mathcal{L}_{\mathrm{ring}}
+
w_{\mathrm{branch}}\mathcal{L}_{\mathrm{branch}}
\right),
\label{eq:soft_rule_loss}
\end{equation}
where auxiliary cross-entropy losses regularize predictions on the corresponding
chemical factors. Detailed definitions and implementation of these auxiliary losses are
provided in Appendix~\ref{app:embed_variants}.

\textbf{Rule-aware logit masking.}
Following DiffMS, at inference time on both NPLIB1 and MassSpecGym we use the
benchmark-provided ground-truth molecular formula, not a separately predicted formula.
The same formula is used by a lightweight constraint controller during autoregressive
decoding and by the post-generation reranker.
Given decoder logits $\mathbf{z}_t\in\mathbb{R}^{|\mathcal{V}|}$ at time step $t$, we
apply a rule-dependent mask $\mathbf{m}_t$ and decode from
\begin{equation}
\tilde{\mathbf{z}}_t
=
\mathbf{z}_t+\mathbf{m}_t,
\qquad
m_{t,j}
\in
\{0,-\infty\},
\label{eq:rule_mask}
\end{equation}
where invalid tokens receive $-\infty$ and are therefore removed from the candidate set.
In our structure-aware decoding design, we use two lightweight chemistry-aware
constraints:
\begin{itemize}
  \item disallow atom tokens whose element does not appear in the target molecular formula;
  \item after a ring or branch token is emitted, enforce the required hexadecimal continuation and forbid isolated hex tokens elsewhere.
\end{itemize}
These constraints directly target frequent failure modes of autoregressive SELFIES
generation: producing chemically impossible elements and mismatching ring/branch control
tokens with their numeric continuations.

\section{Experiments}
\subsection{Experimental Setup}
\label{sec:exp_setup}
Datasets and baselines. The encoder is pretrained on the simulated multimodal
spectroscopy dataset from \citet{alberts2024unraveling}. The decoder is pretrained on
the same 2.8M fingerprint-molecule corpus organized by DiffMS, sampled from DSSTox,
HMDB, COCONUT, and MOSES \citep{bohde2025diffms, wishart2021hmdb,
sorokina2021coconut, polykovskiy2020moses}. For finetuning and evaluation, we follow
the official splits of NPLIB1 and MassSpecGym \citep{bushuiev2024massspecgym}. To prevent data leakage, we remove from both the encoder and decoder pretraining corpora any molecules that overlap with the test sets of NPLIB1 or MassSpecGym. 
We compare against three categories of baselines: (i) naïve autoregressive models
(SMILES/SELFIES Transformers), (ii) standard fingerprint-conditioned methods (Spec2Mol,
MIST+MSNovelist, and MIST+NeuralDecipher), and (iii) recent leading systems
(MADGEN, DiffMS, MS-BART, and FOAM) \citep{goldman2023annotating, stravs2022msnovelist,
bohde2025diffms, han2025msbart, wang2025madgen, manjrekar2026foam}.

\textbf{Metrics.}
We report Top-$k$ metrics with $k\in\{1,10\}$. Our primary identification metric is
{Top-$k$ exact match}, computed by converting each generated molecule to a full
InChIKey and checking whether the ground-truth molecule appears in the top-$k$ set. As
structural-similarity metrics, we report {Top-$k$ maximum Tanimoto similarity}
using RDKit Morgan fingerprints (radius $=2$, length $=4096$) and {Top-$k$
minimum MCES} to quantify graph-level overlap between predictions and targets.

\textbf{Formula information, decoding, and reranking.}
Following DiffMS, all main experiments on both benchmarks use the benchmark-provided
ground-truth molecular formula rather than a separately predicted formula. CoRe-Gen
uses it to mask atom tokens whose elements are absent from the formula, but does not fix
elemental counts or guarantee an exact formula match. Following MS-BART, we generate
100 beam-search candidates per spectrum (beam size $100$)
and rerank them by formula distance
$D(F_1,F_2)=\sum_{a\in\mathcal{A}}|n_a-m_a|$, breaking ties by model-estimated
log-probability. DiffMS instead fixes the heavy-atom inventory before graph generation
and ranks 100 sampled graphs by sampling frequency, so equivalent post-generation
formula reranking is unnecessary. The candidate budget, formula information, and
evaluation metrics are matched; generation-specific ranking is retained for each
paradigm. Appendix~\ref{app:formula_fairness} gives the full comparison and evaluates
MSBuddy-predicted formulas.

\begin{table*}[t]
  \centering
  \caption{Main comparison on NPLIB1 and MassSpecGym. We report Top-1/Top-10 Accuracy ($\uparrow$), MCES ($\downarrow$), and Tanimoto ($\uparrow$). Results marked with $^*$ are reproduced from benchmark or prior reports. $^\dagger$ uses the publicly released DiffMS MassSpecGym encoder checkpoint. Bold indicates the best and underlined indicates the second-best.}
  \label{tab:main_results}
  \small
  \setlength{\tabcolsep}{5pt}
  \renewcommand{\arraystretch}{1.1}
  \resizebox{\textwidth}{!}{%
  \begin{tabular}{lcccccc}
    \toprule
    Model & \multicolumn{3}{c}{Top-1} & \multicolumn{3}{c}{Top-10} \\
    \cmidrule(lr){2-4}\cmidrule(lr){5-7}
    & Accuracy$\uparrow$ & MCES$\downarrow$ & Tanimoto$\uparrow$ & Accuracy$\uparrow$ & MCES$\downarrow$ & Tanimoto$\uparrow$ \\
    \midrule
    \multicolumn{7}{c}{\textbf{NPLIB1}} \\
    \midrule
    Spec2Mol$^*$ & 0.00\% & 27.82 & 0.12 & 0.00\% & 23.13 & 0.16 \\
    MIST + NeuralDecipher$^*$ & 2.32\% & 12.11 & 0.35 & 6.11\% & 9.91 & 0.43 \\
    MIST + MSNovelist$^*$ & 5.40\% & 14.52 & 0.34 & 11.04\% & 10.23 & 0.44 \\
    MADGEN & 2.10\% & 20.56 & 0.22 & 2.39\% & 12.69 & 0.27 \\
    DiffMS & \underline{8.34\%} & 11.95 & 0.35 & \underline{15.44\%} & 9.23 & 0.47 \\
    MS-BART & 7.45\% & \underline{9.66} & \underline{0.44} & 10.99\% & \underline{8.31} & \underline{0.51} \\
    \textbf{CoRe-Gen} & \textbf{21.49\%} & \textbf{5.57} & \textbf{0.57} & \textbf{35.42\%} & \textbf{4.03} & \textbf{0.67} \\
    \midrule
    \multicolumn{7}{c}{\textbf{MassSpecGym}} \\
    \midrule
    SMILES Transformer$^*$ & 0.00\% & 79.39 & 0.03 & 0.00\% & 52.13 & 0.10 \\
    SELFIES Transformer$^*$ & 0.00\% & 38.88 & 0.08 & 0.00\% & 26.87 & 0.13 \\
    Random Generation$^*$ & 0.00\% & 21.11 & 0.08 & 0.00\% & 18.26 & 0.11 \\
    Spec2Mol$^*$ & 0.00\% & 37.76 & 0.12 & 0.00\% & 29.40 & 0.16 \\
    MIST + NeuralDecipher$^*$ & 0.00\% & 33.19 & 0.14 & 0.00\% & 31.89 & 0.16 \\
    MIST + MSNovelist$^*$ & 0.00\% & 45.55 & 0.06 & 0.00\% & 30.13 & 0.15 \\
    MADGEN & 1.31\% & 27.47 & 0.20 & 1.54\% & 16.84 & 0.26 \\
    MS-BART & 1.07\% & 16.47 & 0.23 & 1.11\% & 15.12 & 0.28 \\
    FOAM (DiffMS + PubChem) & 1.50\% & \textbf{12.21} & \underline{0.35} & \underline{10.28\%} & \textbf{6.06} & \textbf{0.53} \\
    DiffMS & \underline{2.30\%} & 18.45 & 0.28 & 4.25\% & 14.73 & 0.39 \\
    \textbf{CoRe-Gen}$^\dagger$ & \textbf{20.02\%} & \underline{16.09} & \textbf{0.40} & \textbf{22.45\%} & \underline{13.90} & \underline{0.47} \\
    \bottomrule
  \end{tabular}%
  }
\end{table*}

\subsection{Main Results}
Before evaluating molecular generation, we first verify that encoder pretraining
substantially improves the predicted fingerprint condition. As shown in
Table~\ref{tab:pretrain_effect}, pretraining reduces BCE by 29.03\% on NPLIB1 and
23.81\% on MassSpecGym, while also improving probability-level cosine similarity. This
confirms that the downstream decoder receives a more usable condition before generation
is evaluated.

Table~\ref{tab:main_results} summarizes overall performance. On NPLIB1, CoRe-Gen
raises Top-1 exact-match accuracy from 8.34\% (DiffMS) to 21.49\% and Top-10 accuracy
from 15.44\% to 35.42\%. Its Top-1/Top-10 MCES reaches 5.57/4.03, while Tanimoto
similarity reaches 0.57/0.67.
Together with Table~\ref{tab:ablation_core}, these gains show that preserving soft
fingerprint confidence and jointly adapting the encoder and decoder substantially
improves both exact recovery and structural similarity.

\begin{wraptable}{r}{0.48\columnwidth}
  \vspace{-1.8em}
  \centering
  \small
  \caption{Encoder pretraining improves fingerprint prediction.}
  \label{tab:pretrain_effect}
  \setlength{\tabcolsep}{4pt}
  \begin{tabular}{lccc}
    \toprule
    Metric & Before & CoRe-Gen & Impr. (\%) \\
    \midrule
    \multicolumn{4}{c}{\textbf{NPLIB1}} \\
    \midrule
    BCE$\downarrow$ & 0.031 & 0.022 & -29.03 \\
    CosSim$\uparrow$ & 0.705 & 0.756 & +7.23 \\
    \midrule
    \multicolumn{4}{c}{\textbf{MassSpecGym}} \\
    \midrule
    BCE$\downarrow$ & 0.063 & 0.048 & -23.81 \\
    CosSim$\uparrow$ & 0.490 & 0.561 & +14.49 \\
    \bottomrule
  \end{tabular}
  \vspace{-1em}
\end{wraptable}

On MassSpecGym, our independently trained MIST encoder did not reproduce the fingerprint
quality of the publicly released DiffMS checkpoint despite following the released code,
configurations, and reported hyperparameters. To align the encoder condition with the
compared DiffMS-based methods, Table~\ref{tab:main_results} therefore uses that public
checkpoint; the CoRe-Gen decoder, data split, formula information, candidate budget,
reranking, and metrics remain unchanged. Under this aligned setting, CoRe-Gen reaches
20.02\%/22.45\% Top-1/Top-10 exact match, 16.09/13.90 MCES, and 0.40/0.47 Tanimoto
similarity. It achieves the best exact-match and Top-1 Tanimoto results, while FOAM
remains strongest on MCES and Top-10 Tanimoto. Appendix~\ref{app:massspecgym_alignment} reports the reproduction details,
independently trained encoder results, and fingerprint quality.

\begin{wraptable}[9]{r}{0.38\columnwidth}
  \vspace{-1.2em}
  \centering
  \scriptsize
  \caption{Inference time per spectrum for 100 candidates on a single RTX 4090 GPU.}
  \label{tab:inference_cost}
  \setlength{\tabcolsep}{3pt}
  \begin{tabular}{@{}lcc@{}}
    \toprule
    Method & Time (s) & Rel. Speedup \\
    \midrule
    DiffMS & $\approx$160 & 1.00$\times$ \\
    MADGEN & $\approx$29 & 5.52$\times$ \\
    CoRe-Gen & $\approx$2 & 80.0$\times$ \\
    \bottomrule
  \end{tabular}
  \vspace{-0.6em}
\end{wraptable}

\begin{figure}[!t]
  \centering
  \includegraphics[width=0.98\textwidth]{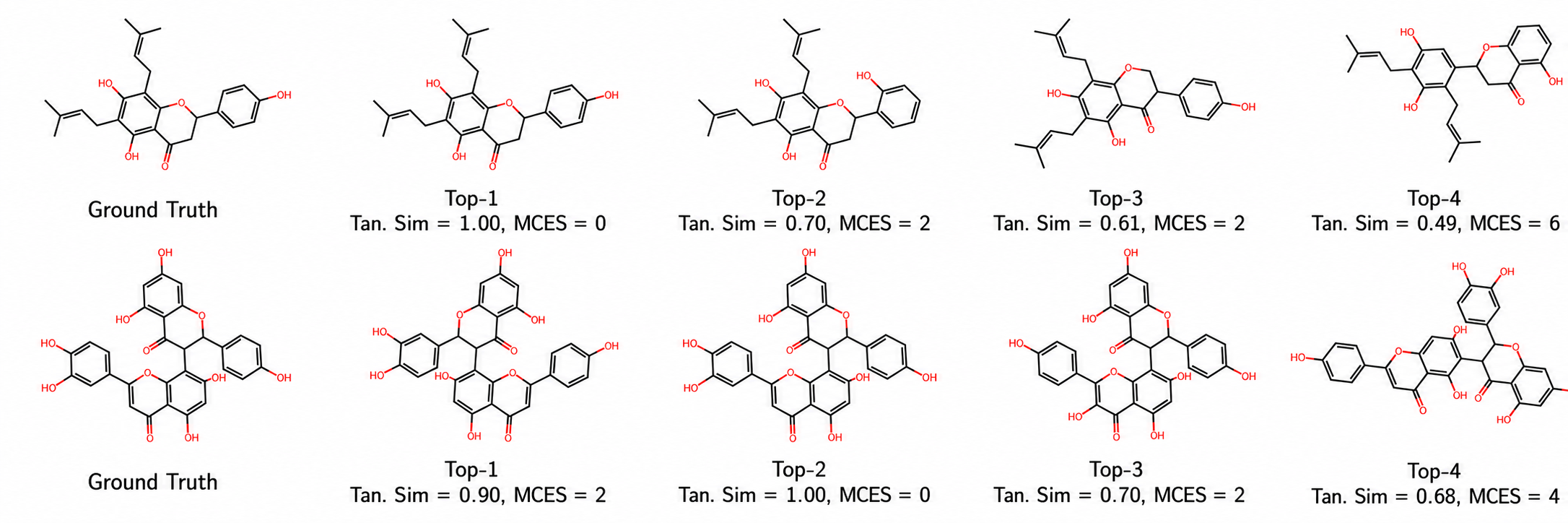}
  \caption{Representative successful generation examples of CoRe-Gen.}
  \label{fig:qual_examples}
\end{figure}

CoRe-Gen is also faster at inference, generating 100 candidates in approximately 2 seconds
per spectrum on a single RTX 4090 GPU---about 80$\times$ faster than DiffMS in our setup.
Figure~\ref{fig:qual_examples} shows representative successful generations, with
additional success and failure cases reported in the appendix.

\subsection{Efficiency and Inference Cost}
\label{sec:efficiency_cost}
At inference time, CoRe-Gen uses beam-search decoding for structure generation. Thanks
to structure-aware compositional embeddings, the effective decoding vocabulary is
compact (43 tokens), which reduces branching complexity per step and improves practical
decoding efficiency while preserving chemical validity constraints.

We systematically evaluate beam width and observe that increasing beam size yields clear
Top-1/Top-10 gains on NPLIB1 while keeping latency practical. As summarized in
Table~\ref{tab:inference_cost}, in a deployment-oriented setting that produces 100
candidates per spectrum on a single RTX 4090 GPU, CoRe-Gen requires approximately 2
seconds per spectrum, compared with approximately 29 seconds for MADGEN and 160
seconds for DiffMS. This corresponds to roughly a 15$\times$ speedup over MADGEN and
an $\sim$80$\times$ speedup over DiffMS, showing that our structure-aware sequence
decoder retains the practical efficiency advantage of autoregressive generation while
remaining competitive in structural recovery.

\setlength{\columnsep}{6pt}
\begin{wraptable}{r}{0.49\columnwidth}
  \vspace{-2em}
  \centering
  \scriptsize
  \caption{Core ablations of CoRe-Gen.}
  \label{tab:ablation_core}
  \setlength{\tabcolsep}{2pt}
  \begin{tabular}{@{}lccc@{}}
    \toprule
    Variant & Top-1 & Top-10 & Top-1 Tan. \\
    \midrule
    Full method & 21.49 & 35.42 & 0.57 \\
    w/o joint finetune & 19.54 & 29.92 & 0.56 \\
    w/o enc. pretrain & 11.46 & 19.80 & 0.48 \\
    w/o corr. pretrain & 14.77 & 23.93 & 0.53 \\
    \midrule
    w/o comp. SELFIES & 18.19 & 26.86 & 0.55 \\
    w/o soft reg. & 19.23 & 29.34 & 0.56 \\
    w/o rule-aware gen. & 19.17 & 28.57 & 0.56 \\
    w/o exploit trio & 17.95 & 26.98 & 0.55 \\
    \midrule
    GT fingerprint & 82.91 & 87.30 & 0.94 \\
    \bottomrule
  \end{tabular}
  \vspace{-2em}
\end{wraptable}
\setlength{\columnsep}{10pt}

\subsection{Ablation Studies}
We conduct ablation studies on NPLIB1. Table~\ref{tab:ablation_core} tests both the
condition-centric components and the differentiable interface. Removing joint finetuning
reduces Top-1/Top-10 accuracy from 21.49\%/35.42\% to 19.54\%/29.92\%, with Top-1
Tanimoto decreasing from 0.57 to 0.56. The larger Top-10 gain indicates that end-to-end
adaptation improves candidate-set coverage and ranking rather than only the first output.

The largest contribution comes from encoder pretraining. Removing it drops Top-1
accuracy from 21.49\% to 11.46\% and Top-10 accuracy from 35.42\% to 19.80\%, showing
that synthetic spectrum-to-fingerprint supervision is critical for learning a reliable
condition signal before adaptation on scarce real paired data. This large gain is
consistent with the fingerprint-prediction improvements reported in Table~\ref{tab:pretrain_effect}, where pretraining substantially improves the predicted
fingerprints used by the downstream generator.

Corruption-aware decoder pretraining is also important. Without it, Top-1 and Top-10
accuracy decrease to 14.77\% and 23.93\%, respectively, confirming that
the decoder must be trained under noisy predicted-fingerprint conditions rather than
only clean oracle fingerprints.

Compositional SELFIES embedding and its lightweight refinements provide smaller but
consistent additional gains. Removing the compositional SELFIES embedding lowers
Top-1/Top-10 accuracy to 18.19\%/26.86\%, while removing soft regularization or
rule-aware generation produces more modest declines. These results indicate that
compositional tokenization supplies the main local chemical inductive bias, with
auxiliary supervision and decoding rules serving as useful refinements on top of
improved and better-matched conditions. When we remove the full exploit module
('w/o exploit trio'), Top-1 accuracy further drops to 17.95\%, confirming that the
combined effect of compositional embedding, soft regularization, and rule-aware
generation is meaningful under imperfect fingerprint conditioning.

Finally, the 'GT fingerprint' row (Top-1 82.91\%, Top-10 87.30\%) indicates substantial
headroom beyond all predicted-fingerprint settings, suggesting that the dominant
bottleneck is still the encoder's ability to predict highly accurate fingerprints.

\section{Conclusion}
We presented a condition-centric framework for fingerprint-conditioned
spectrum-to-structure generation that treats imperfect predicted fingerprints as the
central bottleneck of practical deployment. CoRe-Gen improves the predicted condition
through synthetic-spectrum encoder pretraining, matches deployment-time corruption
through corruption-aware decoder pretraining, and replaces the discrete fingerprint
handoff with a differentiable memory bridge for joint encoder--decoder adaptation. It
then exploits residual condition noise through structure-aware autoregressive decoding.
Across standard benchmarks, this
design surpasses leading compared methods on NPLIB1 and, under an encoder-aligned
evaluation, on MassSpecGym exact-match accuracy, while preserving the
practical efficiency advantage of autoregressive generation.
More broadly, robust spectrum-to-structure generation should model
intermediate-condition uncertainty through stronger encoders, condition-aligned
pretraining, and deployment-robust decoding.

\subsection*{AI use statement}
In preparing this manuscript, we used generative AI tools to assist with language
editing, document restructuring, \LaTeX{} formatting, reference-format checking, and
migration to the ICLR 2027 template. The authors reviewed the AI-assisted edits against
the underlying text, numerical results, citations, tables, and figures. The authors take
responsibility for the final content of this work, including text or artifacts produced
with the aid of generative AI.

\subsection*{Ethics statement}
CoRe-Gen produces ranked structural hypotheses rather than definitive molecular
identifications. Its outputs should be validated with reference spectra or orthogonal
experimental evidence before being used in scientific, clinical, forensic, or safety-
critical decisions. Like other structure-elucidation systems, it also has potential
dual-use implications for identifying harmful or illicit compounds. We discuss these
limitations and responsible-use considerations further in
Appendix~\ref{app:broader_impacts}.

\subsection*{Reproducibility statement}
The appendix documents the datasets and split rules, molecular and spectral
representations, training schedule and hyperparameters, decoding and reranking
protocols, encoder-alignment procedure, baseline-comparison settings, and variation
across random seeds. It also reports robustness analyses for predicted formulas,
training-set coverage, scaffold recovery, and library depletion. Together with the
equations and algorithmic descriptions in the main paper, these details specify the
reported experimental pipeline and evaluation procedure.

\bibliography{coregen_arxiv}

@article{alberts2024unraveling,
  title={{Unraveling molecular structure: A multimodal spectroscopic dataset for chemistry}},
  author={Alberts, Marvin and Schilter, Oliver and Zipoli, Federico and Hartrampf, Nina and Laino, Teodoro},
  journal={Advances in neural information processing systems},
  volume={37},
  pages={125780--125808},
  year={2024}
}

@article{allen2014cfmid,
  title={{CFM-ID: a web server for annotation, spectrum prediction and metabolite identification from tandem mass spectra}},
  author={Allen, Felicity and Pon, Allison and Wilson, Michael and Greiner, Russ and Wishart, David},
  journal={Nucleic acids research},
  volume={42},
  number={W1},
  pages={W94--W99},
  year={2014},
  publisher={Oxford University Press}
}

@article{cao2021moldiscovery,
  title={{MolDiscovery: learning mass spectrometry fragmentation of small molecules}},
  author={Cao, Liu and Guler, Mustafa and Tagirdzhanov, Azat and Lee, Yi-Yuan and Gurevich, Alexey and Mohimani, Hosein},
  journal={Nature communications},
  volume={12},
  number={1},
  pages={3718},
  year={2021},
  publisher={Nature Publishing Group UK London}
}

@article{butler2023ms2mol,
  title={{MS2Mol: A transformer model for illuminating dark chemical space from mass spectra}},
  author={Butler, Thomas and Frandsen, Abraham and Lightheart, Rose and Bargh, Brian and Kerby, Thomas and West, Kiana and Davison, Joseph and Taylor, James and Krettler, Christoph and Bollerman, TJ and others},
  journal={ChemRxiv},
  year={2023},
  publisher={ChemRxiv},
  doi={10.26434/chemrxiv-2023-vsmpx-v4}
}

@article{bohde2025diffms,
  title={{DiffMS: Diffusion generation of molecules conditioned on mass spectra}},
  author={Bohde, Montgomery and Manjrekar, Mrunali and Wang, Runzhong and Ji, Shuiwang and Coley, Connor W},
  journal={arXiv preprint arXiv:2502.09571},
  year={2025}
}

@article{brouard2016iokr,
  title={{Fast metabolite identification with input output kernel regression}},
  author={Brouard, C{\'e}line and Shen, Huibin and D{\"u}hrkop, Kai and d'Alch{\'e}-Buc, Florence and B{\"o}cker, Sebastian and Rousu, Juho},
  journal={Bioinformatics},
  volume={32},
  number={12},
  pages={i28--i36},
  year={2016},
  publisher={Oxford University Press}
}

@article{bushuiev2024massspecgym,
  title={{MassSpecGym: A benchmark for the discovery and identification of molecules}},
  author={Bushuiev, Roman and Bushuiev, Anton and de Jonge, Niek F and Young, Adamo and Kretschmer, Fleming and Samusevich, Raman and Heirman, Janne and Wang, Fei and Zhang, Luke and D{\"u}hrkop, Kai and others},
  journal={Advances in Neural Information Processing Systems},
  volume={37},
  pages={110010--110027},
  year={2024}
}

@article{dejonge2023ms2query,
  title={{MS2Query: reliable and scalable MS2 mass spectra-based analogue search}},
  author={de Jonge, Niek F and Louwen, Joris JR and Chekmeneva, Elena and Camuzeaux, Stephane and Vermeir, Femke J and Jansen, Robert S and Huber, Florian and van der Hooft, Justin JJ},
  journal={Nature Communications},
  volume={14},
  number={1},
  pages={1752},
  year={2023},
  publisher={Nature Publishing Group UK London}
}

@article{duhrkop2015searching,
  title={{Searching molecular structure databases with tandem mass spectra using CSI: FingerID}},
  author={D{\"u}hrkop, Kai and Shen, Huibin and Meusel, Marvin and Rousu, Juho and B{\"o}cker, Sebastian},
  journal={Proceedings of the National Academy of Sciences},
  volume={112},
  number={41},
  pages={12580--12585},
  year={2015},
  publisher={National Academy of Sciences}
}

@article{duhrkop2019sirius,
  title={{SIRIUS 4: a rapid tool for turning tandem mass spectra into metabolite structure information}},
  author={D{\"u}hrkop, Kai and Fleischauer, Markus and Ludwig, Marcus and Aksenov, Alexander A and Melnik, Alexey V and Meusel, Marvin and Dorrestein, Pieter C and Rousu, Juho and B{\"o}cker, Sebastian},
  journal={Nature methods},
  volume={16},
  number={4},
  pages={299--302},
  year={2019},
  publisher={Nature Publishing Group US New York}
}

@article{goldman2023annotating,
  title={{MIST-CF: Chemical formula inference from tandem mass spectra}},
  author={Goldman, Samuel and Xin, Jiayi and Provenzano, Joules and Coley, Connor W},
  journal={Journal of chemical information and modeling},
  volume={64},
  number={7},
  pages={2421--2431},
  year={2024},
  publisher={ACS Publications}
}

@article{goldman2023iceberg,
  title={{Generating molecular fragmentation graphs with autoregressive neural networks}},
  author={Goldman, Samuel and Li, Janet and Coley, Connor W},
  journal={Analytical Chemistry},
  volume={96},
  number={8},
  pages={3419--3428},
  year={2024},
  doi={10.1021/acs.analchem.3c04654}
}

@article{goldman2023scarf,
  title={{Prefix-tree decoding for predicting mass spectra from molecules}},
  author={Goldman, Samuel and Bradshaw, John and Xin, Jiayi and Coley, Connor},
  journal={Advances in neural information processing systems},
  volume={36},
  pages={48548--48572},
  year={2023}
}

@article{guijas2018metlin,
  title={{METLIN: a technology platform for identifying knowns and unknowns}},
  author={Guijas, Carlos and Montenegro-Burke, J Rafael and Domingo-Almenara, Xavier and Palermo, Amelia and Warth, Benedikt and Hermann, Gerrit and Koellensperger, Gunda and Huan, Tao and Uritboonthai, Winnie and Aisporna, Aries E and others},
  journal={Analytical chemistry},
  volume={90},
  number={5},
  pages={3156--3164},
  year={2018},
  publisher={ACS Publications}
}

@inproceedings{han2025msbart,
  title={{MS-BART: Unified Modeling of Mass Spectra and Molecules for Structure Elucidation}},
  author={Han, Yang and Wang, Pengyu and Yu, Kai and Chen, Xin and Chen, Lu},
  booktitle={Advances in Neural Information Processing Systems},
  volume={38},
  pages={111459--111481},
  year={2025},
  doi={10.52202/085713-3721}
}

@article{huber2021ms2deepscore,
  title={{MS2DeepScore: a novel deep learning similarity measure to compare tandem mass spectra}},
  author={Huber, Florian and van der Burg, Sven and van der Hooft, Justin JJ and Ridder, Lars},
  journal={Journal of cheminformatics},
  volume={13},
  number={1},
  pages={84},
  year={2021},
  publisher={Springer}
}

@article{horai2010massbank,
  title={{MassBank: a public repository for sharing mass spectral data for life sciences}},
  author={Horai, Hisayuki and Arita, Masanori and Kanaya, Shigehiko and Nihei, Yoshito and Ikeda, Tasuku and Suwa, Kazuhiro and Ojima, Yuya and Tanaka, Kenichi and Tanaka, Satoshi and Aoshima, Ken and others},
  journal={Journal of mass spectrometry},
  volume={45},
  number={7},
  pages={703--714},
  year={2010},
  publisher={Wiley Online Library}
}

@article{ji2020deepei,
  title={{Predicting a molecular fingerprint from an electron ionization mass spectrum with deep neural networks}},
  author={Ji, Hongchao and Deng, Hanzi and Lu, Hongmei and Zhang, Zhimin},
  journal={Analytical chemistry},
  volume={92},
  number={13},
  pages={8649--8653},
  year={2020},
  publisher={ACS Publications}
}

@article{krenn2020selfies,
  title={{Self-referencing embedded strings (SELFIES): A 100\% robust molecular string representation}},
  author={Krenn, Mario and H{\"a}se, Florian and Nigam, AkshatKumar and Friederich, Pascal and Aspuru-Guzik, Alan},
  journal={Machine Learning: Science and Technology},
  volume={1},
  number={4},
  pages={045024},
  year={2020},
  doi={10.1088/2632-2153/aba947}
}

@article{li2021spectralentropy,
  title={{Spectral entropy outperforms MS/MS dot product similarity for small-molecule compound identification}},
  author={Li, Yuanyue and Kind, Tobias and Folz, Jacob and Vaniya, Arpana and Mehta, Sajjan Singh and Fiehn, Oliver},
  journal={Nature Methods},
  volume={18},
  number={12},
  pages={1524--1531},
  year={2021},
  publisher={Nature Publishing Group US New York}
}

@article{litsa2023spec2mol,
  title={{An end-to-end deep learning framework for translating mass spectra to de-novo molecules}},
  author={Litsa, Eleni E and Chenthamarakshan, Vijil and Das, Payel and Kavraki, Lydia E},
  journal={Communications Chemistry},
  volume={6},
  number={1},
  pages={132},
  year={2023},
  publisher={Nature Publishing Group UK London}
}

@inproceedings{liu2023ms2smiles,
  title={{De novo molecular structure generation from mass spectra}},
  author={Liu, Yanmin and Zhang, Xuan and Zhao, Wei and Zhu, Daming and Cui, Xuefeng},
  booktitle={2023 IEEE International Conference on Bioinformatics and Biomedicine (BIBM)},
  pages={373--378},
  year={2023},
  organization={IEEE}
}

@article{manjrekar2026foam,
  title={{Generative structural elucidation from mass spectra as an iterative optimization problem}},
  author={Manjrekar, Mrunali and Wang, Runzhong and Goldman, Samuel and Fromer, Jenna C and Coley, Connor W},
  journal={arXiv preprint arXiv:2602.07709},
  year={2026}
}

@article{polykovskiy2020moses,
  title={{Molecular sets (MOSES): a benchmarking platform for molecular generation models}},
  author={Polykovskiy, Daniil and Zhebrak, Alexander and Sanchez-Lengeling, Benjamin and Golovanov, Sergey and Tatanov, Oktai and Belyaev, Stanislav and Kurbanov, Rauf and Artamonov, Aleksey and Aladinskiy, Vladimir and Veselov, Mark and others},
  journal={Frontiers in pharmacology},
  volume={11},
  pages={565644},
  year={2020},
  publisher={Frontiers Media SA}
}

@article{ruttkies2016metfrag,
  title={{MetFrag relaunched: incorporating strategies beyond in silico fragmentation}},
  author={Ruttkies, Christoph and Schymanski, Emma L and Wolf, Sebastian and Hollender, Juliane and Neumann, Steffen},
  journal={Journal of cheminformatics},
  volume={8},
  number={1},
  pages={3},
  year={2016},
  publisher={Springer}
}

@article{scheubert2013computational,
  title={{Computational mass spectrometry for small molecules}},
  author={Scheubert, Kerstin and Hufsky, Franziska and B{\"o}cker, Sebastian},
  journal={Journal of cheminformatics},
  volume={5},
  number={1},
  pages={12},
  year={2013},
  publisher={Springer}
}

@article{schymanski2013casmi,
  title={{The critical assessment of small molecule identification (CASMI): challenges and solutions}},
  author={Schymanski, Emma L and Neumann, Steffen},
  journal={Metabolites},
  volume={3},
  number={3},
  pages={517--538},
  year={2013},
  publisher={MDPI}
}

@article{schymanski2017casmi2016,
  title={{Critical assessment of small molecule identification 2016: automated methods}},
  author={Schymanski, Emma L and Ruttkies, Christoph and Krauss, Martin and Brouard, C{\'e}line and Kind, Tobias and D{\"u}hrkop, Kai and Allen, Felicity and Vaniya, Arpana and Verdegem, Dries and B{\"o}cker, Sebastian and others},
  journal={Journal of cheminformatics},
  volume={9},
  number={1},
  pages={22},
  year={2017},
  publisher={Springer}
}

@article{sorokina2021coconut,
  title={{COCONUT online: collection of open natural products database}},
  author={Sorokina, Maria and Merseburger, Peter and Rajan, Kohulan and Yirik, Mehmet Aziz and Steinbeck, Christoph},
  journal={Journal of Cheminformatics},
  volume={13},
  number={1},
  pages={2},
  year={2021},
  publisher={Springer}
}

@article{stein2012mass,
  title={{Mass spectral reference libraries: an ever-expanding resource for chemical identification}},
  author={Stein, Stephen},
  journal={Analytical Chemistry},
  volume={84},
  number={17},
  pages={7274--7282},
  year={2012},
  doi={10.1021/ac301205z}
}

@article{stravs2022msnovelist,
  title={{MSNovelist: de novo structure generation from mass spectra}},
  author={Stravs, Michael A and D{\"u}hrkop, Kai and B{\"o}cker, Sebastian and Zamboni, Nicola},
  journal={Nature Methods},
  volume={19},
  number={7},
  pages={865--870},
  year={2022},
  publisher={Nature Publishing Group US New York}
}

@article{ucak2020neuraldecipher,
  title={{Neuraldecipher--reverse-engineering extended-connectivity fingerprints (ECFPs) to their molecular structures}},
  author={Le, Tuan and Winter, Robin and No{\'e}, Frank and Clevert, Djork-Arn{\'e}},
  journal={Chemical science},
  volume={11},
  number={38},
  pages={10378--10389},
  year={2020},
  publisher={The Royal Society of Chemistry}
}

@article{ucak2023reconstruction,
  title={{Reconstruction of lossless molecular representations from fingerprints}},
  author={Ucak, Umit V and Ashyrmamatov, Islambek and Lee, Juyong},
  journal={Journal of cheminformatics},
  volume={15},
  number={1},
  pages={26},
  year={2023},
  publisher={Springer}
}

@article{wang2016sharing,
  title={{Sharing and community curation of mass spectrometry data with Global Natural Products Social Molecular Networking}},
  author={Wang, Mingxun and Carver, Jeremy J and Phelan, Vanessa V and Sanchez, Laura M and Garg, Neha and Peng, Yao and Nguyen, Don Duy and Watrous, Jeramie and Kapono, Clifford A and Luzzatto-Knaan, Tal and others},
  journal={Nature biotechnology},
  volume={34},
  number={8},
  pages={828--837},
  year={2016},
  publisher={Nature Publishing Group US New York}
}

@article{wang2022cfmid4,
  title={{CFM-ID 4.0--a web server for accurate MS-based metabolite identification}},
  author={Wang, Fei and Allen, Dana and Tian, Siyang and Oler, Eponine and Gautam, Vasuk and Greiner, Russell and Metz, Thomas O and Wishart, David S},
  journal={Nucleic acids research},
  volume={50},
  number={W1},
  pages={W165--W174},
  year={2022},
  publisher={Oxford University Press}
}

@inproceedings{wang2025madgen,
  title={{MADGEN: Mass-spec attends to de novo molecular generation}},
  author={Wang, Yinkai and Chen, Xiaohui and Liu, Liping and Hassoun, Soha},
  booktitle={International Conference on Learning Representations},
  year={2025},
  url={https://openreview.net/forum?id=78tc3EiUrN}
}

@article{wei2019neims,
  title={{Rapid prediction of electron--ionization mass spectrometry using neural networks}},
  author={Wei, Jennifer N and Belanger, David and Adams, Ryan P and Sculley, D},
  journal={ACS central science},
  volume={5},
  number={4},
  pages={700},
  year={2019}
}

@article{wishart2021hmdb,
  title={{HMDB 5.0: the human metabolome database for 2022}},
  author={Wishart, David S and Guo, AnChi and Oler, Eponine and Wang, Fei and Anjum, Afia and Peters, Harrison and Dizon, Raynard and Sayeeda, Zinat and Tian, Siyang and Lee, Brian L and others},
  journal={Nucleic acids research},
  volume={50},
  number={D1},
  pages={D622--D631},
  year={2022},
  publisher={Oxford University Press}
}

@article{wolf2010metfrag,
  title={{In silico fragmentation for computer assisted identification of metabolite mass spectra}},
  author={Wolf, Sebastian and Schmidt, Stephan and M{\"u}ller-Hannemann, Matthias and Neumann, Steffen},
  journal={BMC bioinformatics},
  volume={11},
  number={1},
  pages={148},
  year={2010},
  publisher={Springer}
}

@article{spec2vec,
  title={{Spec2Vec: Improved mass spectral similarity scoring through learning of structural relationships}},
  author={Huber, Florian and Ridder, Lars and Verhoeven, Stefan and Spaaks, Jurriaan H and Diblen, Faruk and Rogers, Simon and Van Der Hooft, Justin JJ},
  journal={PLoS computational biology},
  volume={17},
  number={2},
  pages={e1008724},
  year={2021},
  publisher={Public Library of Science San Francisco, CA USA}
}

@article{young2024massformer,
  title={{Tandem mass spectrum prediction for small molecules using graph transformers}},
  author={Young, Adamo and R{\"o}st, Hannes and Wang, Bo},
  journal={Nature Machine Intelligence},
  volume={6},
  number={4},
  pages={404--416},
  year={2024},
  publisher={Nature Publishing Group UK London}
}

@article{xing2023buddy,
  title={{BUDDY: molecular formula discovery via bottom-up MS/MS interrogation}},
  author={Xing, Shipei and Shen, Sam and Xu, Banghua and Li, Xiaoxiao and Huan, Tao},
  journal={Nature Methods},
  volume={20},
  number={6},
  pages={881--890},
  year={2023},
  publisher={Nature Publishing Group US New York}
}

@article{zhang2024msbert,
  title={{MSBERT: embedding tandem mass spectra into chemically rational space by mask learning and contrastive learning}},
  author={Zhang, Hailiang and Yang, Qiong and Xie, Ting and Wang, Yue and Zhang, Zhimin and Lu, Hongmei},
  journal={Analytical Chemistry},
  volume={96},
  number={42},
  pages={16599--16608},
  year={2024},
  publisher={ACS Publications}
}
\bibliographystyle{iclr2027_conference}

\appendix
\section{Experimental Details}
\label{app:exp_details}
\subsection{Datasets, Pretraining Corpora, and Split Rules}
\label{app:datasets}
The encoder is pretrained on large-scale simulated spectra with Morgan-4096 fingerprint
targets. The decoder is pretrained on a large fingerprint--molecule corpus constructed
independently of the downstream spectrum-conditioned benchmarks and kept aligned with
the DiffMS setup. The final encoder and decoder adaptation uses experimental spectra
and the corresponding encoder-predicted fingerprints.

The encoder-pretraining and decoder-pretraining splits follow the corresponding
original data sources, and we strictly remove from the pretraining training sets any
molecules that overlap with the test sets of the two finetuning benchmarks, NPLIB1 and
MassSpecGym.

We use different pretraining corpora for two reasons. First, decoder pretraining follows
the 2.8M fingerprint--molecule corpus organized by DiffMS, which keeps the decoder-side
setup aligned with the strongest baseline. Second, the current bottleneck is the
spectrum encoder rather than the fingerprint-to-structure decoder: the decoder corpus is
already several times larger than the simulated spectroscopic corpus, whereas the
additional data needed by the encoder must contain spectra. Since the DiffMS
decoder-pretraining corpus has no spectra, encoder pretraining relies on the simulated
spectrum--molecule corpus instead.

For downstream evaluation, we keep the dataset partitions aligned with DiffMS. In
particular, NPLIB1 follows the NPLIB1 split used in DiffMS, and MassSpecGym
follows the same split protocol as the DiffMS benchmark export. Thus, all reported
results on NPLIB1 and MassSpecGym are obtained under the same train/validation/test
partitions as DiffMS.

For final joint finetuning, the bridge receives the encoder's soft fingerprint
probabilities together with the target modified SELFIES sequence. No thresholding or
active-bit index selection is applied, so the sequence loss can update the spectrum
encoder through the differentiable bridge.

\subsection{Preprocessing and Representation Details}
\label{app:preprocess}
Across encoder training, the prediction target is the Morgan 4096-bit fingerprint
(\texttt{morgan4096}). The encoder uses the standard MIST spectrum inputs together with
subformula and MAGMa side information when available. During encoder pretraining and
finetuning, spectra are augmented by peak removal and intensity perturbation. The
simulated pretraining configuration uses augmentation probability $0.5$, removal
probability $0.5$, intensity perturbation probability $0.1$, and exponential removal
weights. The NPLIB1 finetuning configuration uses the same augmentation strengths,
while the MassSpecGym finetuning configuration uses a milder setup with augmentation
probability $0.1$ and removal probability $0.1$.

For structure generation, target molecules are represented as modified SELFIES strings
and tokenized into a hexadecimal-compatible modified SELFIES vocabulary. Unless
otherwise noted, the decoder uses our structure-aware compositional embedding, which
factorizes token identity into chemical attributes such as part type, element, bond, and
ring/branch indicators, while omitting charge- and explicit-H-specific factors to avoid
unnecessary sparsity. The
maximum target length is set to 256 during training and increased only for generation
evaluation when needed. In the main experiments, both benchmarks use the
benchmark-provided ground-truth molecular formula. It masks only atom tokens whose
elements are absent from the formula during decoding and is also used for
post-generation formula-distance reranking; elemental counts are not fixed.

\subsection{Training Configuration}
\label{app:training_config}
The encoder backbone follows the MIST architecture with hidden size 512, 8 attention
heads, 3 peak-attention layers, and 5 refinement layers. The decoder side uses a
6-layer autoregressive Transformer with hidden size 512, 8 heads, feed-forward
dimension 2048, and our structure-aware compositional token embedding. The source
interface is our fingerprint-to-memory bridge: it retains all 4096 soft bit values,
folds them into 64 modulo chunks, and contextualizes the resulting memory sequence with
6 Transformer layers of hidden size 512, 8 heads, and RoPE positional encoding.

\paragraph{Encoder pretraining.}
The encoder is pretrained on simulated spectra with batch size 768 over 3 GPUs, learning
rate $8\times 10^{-4}$, weight decay $10^{-7}$, and up to 100 epochs. We use spectra
dropout 0.1, iterative fingerprint prediction with \texttt{iterative-loss-weight}
$=0.4$, cosine loss, and MAGMa auxiliary supervision with coefficient 8. Training uses a
scheduler with decay fraction 0.9 and early stopping on validation loss with patience 20.

\paragraph{Encoder finetuning.}
Encoder finetuning initializes from the simulated-pretrained checkpoint and adapts it on
experimental spectra. On the NPLIB1 pipeline, we use batch size 128 on 2 GPUs, learning
rate $8\times 10^{-5}$, weight decay $10^{-7}$, cosine loss, and up to 300 epochs. On
the MassSpecGym pipeline, we use batch size 192 on 3 GPUs, learning rate $3\times
10^{-5}$, warmup 1000 steps, BCE loss, and up to 600 epochs. The NPLIB1 setting uses
\texttt{frac-orig}=0.6 forward augmentation, while the MassSpecGym setting additionally
uses bit-frequency reweighting for calibration.

\paragraph{Decoder pretraining.}
The fingerprint-to-structure decoder is pretrained with batch size 448 on 2 GPUs,
learning rate $3\times 10^{-4}$, weight decay $2\times 10^{-2}$, warmup 1000, gradient
clipping 1.0, dropout 0.15, and 40 training epochs. We optimize the autoregressive
cross-entropy together with the structure-aware auxiliary sentence loss described in
Section~\ref{app:embed_variants}, using total auxiliary weight 0.1, with factor weights
1.0 for element prediction and 0.2 each for bond, ring, and branch prediction.
Corruption-aware pretraining uses a phased schedule consisting of 5 clean epochs, 5 ramp
epochs, and a full-corruption phase thereafter. The corruption probability ramps from
0.3 to its final value of 0.6. Whenever corruption is applied, the bit-swap budget is
sampled from the clipped-Poisson distribution in
Eq.~\eqref{eq:corruption_budget}, with $\lambda=k_{\max}/2$ and final
$k_{\max}=24$, followed by the bucket-weighted selection in
Eq.~\eqref{eq:corruption_sampling}.

\paragraph{Joint encoder--decoder finetuning.}
The final stage initializes the pretrained spectrum encoder and corruption-aware
decoder, then jointly updates the encoder, fingerprint-to-memory bridge, and decoder.
We train with batch size 672 on 3 GPUs, learning rate $10^{-5}$, weight decay
$4\times 10^{-7}$, dropout 0.1, and 20 epochs. The objective combines autoregressive
cross-entropy, the fingerprint-prediction loss, and the auxiliary structural loss. Model
selection uses decoder cross-entropy.

\subsection{Decoding and Reranking Protocol}
\label{app:decode_rerank}
In all main experiments, CoRe-Gen uses the benchmark-provided ground-truth molecular
formula, generates 100 candidates per spectrum with beam size 100, and reranks them by
molecular-formula distance, following MS-BART. Soft fingerprint probabilities are fed
directly to the bridge without thresholding; ties under formula-distance reranking are
broken by model-estimated log-probability. The predicted-formula experiment in
Appendix~\ref{app:formula_robustness} is the only reported setting that replaces the
ground-truth formula with formulas inferred from the spectrum.

The decoder evaluation scripts support both greedy and beam decoding, but all reported
top-$k$ generation metrics use beam search. During decoder evaluation, the active
finetuning script enables hexadecimal-token normalization, batched beam search, and the
constraint controller during generation. The generation configuration uses
\texttt{beam\_size}$=100$, \texttt{beam\_num}$=100$, \texttt{topk}$=10$, and evaluates on
the test split. This setup matches the protocol used for our comparisons against DiffMS
and other baselines.

\section{MassSpecGym Encoder Alignment}
\label{app:massspecgym_alignment}
CoRe-Gen is a two-stage framework, so generation quality depends directly on the
fingerprint predicted by the MIST encoder. Synthetic-spectrum pretraining improves our
independently trained MassSpecGym encoder on all three fingerprint metrics in
Table~\ref{tab:msg_encoder_quality}, and it also outperforms the MS-BART encoder.
However, it remains weaker than the publicly released DiffMS checkpoint.

\begin{table}[t]
  \centering
  \caption{Fingerprint prediction on MassSpecGym.}
  \label{tab:msg_encoder_quality}
  \small
  \setlength{\tabcolsep}{4pt}
  \begin{tabular}{lccc}
    \toprule
    Encoder & BCE$\downarrow$ & Cosine$\uparrow$ & Tanimoto$\uparrow$ \\
    \midrule
    CoRe-Gen before pretraining & 0.063 & 0.490 & 0.251 \\
    CoRe-Gen & 0.048 & 0.561 & 0.322 \\
    MS-BART & 0.079 & 0.470 & 0.213 \\
    DiffMS released checkpoint & \textbf{0.044} & \textbf{0.623} & \textbf{0.464} \\
    \bottomrule
  \end{tabular}
\end{table}

We attempted to reproduce the released DiffMS checkpoint using the authors' publicly
released code, configurations, and reported hyperparameters, but our independently
trained encoder still did not reach the released checkpoint's fingerprint quality. We
contacted the DiffMS author team for clarification but did not receive a response before
submission. We therefore isolated this unresolved discrepancy in a controlled
encoder-alignment experiment. Because the independently trained encoder gives
CoRe-Gen and DiffMS conditions of different quality, all encoder-aligned results use the
released DiffMS checkpoint; the data split, formula information, candidate budget,
reranking, and metrics remain unchanged.

\begin{table*}[t]
  \centering
  \caption{Effect of encoder alignment on CoRe-Gen. Only the encoder checkpoint changes between rows.}
  \label{tab:msg_encoder_alignment}
  \small
  \setlength{\tabcolsep}{5pt}
  \begin{tabular}{lcccccc}
    \toprule
    & \multicolumn{3}{c}{Top-1} & \multicolumn{3}{c}{Top-10} \\
    \cmidrule(lr){2-4}\cmidrule(lr){5-7}
    Encoder & Accuracy$\uparrow$ & MCES$\downarrow$ & Tanimoto$\uparrow$ & Accuracy$\uparrow$ & MCES$\downarrow$ & Tanimoto$\uparrow$ \\
    \midrule
    Independently trained & 1.23\% & 15.19 & 0.26 & 2.67\% & 13.69 & 0.32 \\
    DiffMS released checkpoint & 20.02\% & 16.09 & 0.40 & 22.45\% & 13.90 & 0.47 \\
    \bottomrule
  \end{tabular}
\end{table*}

In this controlled comparison, changing only the encoder checkpoint raises Top-1/Top-10
exact match from 1.23\%/2.67\% to 20.02\%/22.45\%. Thus, the weak independently trained
result is substantially affected by encoder quality and should not be interpreted solely
as a decoder limitation. The released-checkpoint row is the encoder-aligned CoRe-Gen
setting reported in Table~\ref{tab:main_results}.

\section{Formula Information and Baseline Fairness}
\label{app:formula_fairness}
For both NPLIB1 and MassSpecGym, the main results use the benchmark-provided
ground-truth molecular formula, not a separately predicted formula. All
formula-conditioned methods receive the same formula, but use it according to their
original generation paradigm. DiffMS fixes the number and types of heavy atoms before
graph generation and mainly predicts connectivity. CoRe-Gen only disallows atom tokens
whose elements are absent from the formula; it neither fixes elemental counts nor
guarantees an exact formula match. Following MS-BART, CoRe-Gen then reranks 100
beam-search candidates first by formula distance and then by model probability. It
therefore receives no additional formula information relative to the principal
formula-conditioned baselines.

\begin{table*}[t]
  \centering
  \caption{Evaluation controls and paradigm-specific choices.}
  \label{tab:baseline_fairness}
  \small
  \resizebox{\textwidth}{!}{%
  \begin{tabular}{llll}
    \toprule
    Setting & DiffMS & CoRe-Gen / MS-BART & Standardized? \\
    \midrule
    Candidate budget & 100 sampled graphs & 100 beam-search candidates & Yes \\
    Molecular formula & Benchmark-provided formula & Benchmark-provided formula & Yes \\
    Evaluation metrics & Exact match, MCES, Tanimoto & Exact match, MCES, Tanimoto & Yes \\
    Candidate generation & Graph-diffusion sampling & Autoregressive beam search & No; paradigm-specific \\
    Duplicate handling / ranking & Sampling frequency & Formula distance, then model probability & No; paradigm-specific \\
    \bottomrule
  \end{tabular}%
  }
\end{table*}

DiffMS ranks duplicate structures by sampling frequency, whereas CoRe-Gen retains the
MS-BART ranking rule. Imposing identical duplicate handling would change the methods'
native candidate construction. We therefore standardize the candidate budget, formula
information, and evaluation metrics while retaining each model's original
generation-specific procedure.

\subsection{Robustness to Predicted Molecular Formulas}
\label{app:formula_robustness}
This auxiliary experiment evaluates CoRe-Gen without ground-truth formulas. We use the
Top-5 formulas predicted
directly from each spectrum by MSBuddy~\citep{xing2023buddy} rather than manually perturbing the reference
formula. MSBuddy includes the correct formula much less often on MassSpecGym than on
NPLIB1 (Table~\ref{tab:msbuddy_formula_accuracy}). Candidates are reranked by their
minimum weighted formula distance to these five predictions, with ties broken by
decoder log-probability.

\begin{table}[t]
  \centering
  \caption{MSBuddy molecular-formula prediction accuracy.}
  \label{tab:msbuddy_formula_accuracy}
  \small
  \begin{tabular}{lcc}
    \toprule
    Dataset & Top-1 accuracy & Top-5 accuracy \\
    \midrule
    NPLIB1 & 82.42\% & 84.86\% \\
    MassSpecGym & 49.93\% & 57.69\% \\
    \bottomrule
  \end{tabular}
\end{table}

\begin{table*}[t]
  \centering
  \caption{CoRe-Gen using MSBuddy-predicted formulas instead of benchmark-provided formulas.}
  \label{tab:predicted_formula_results}
  \small
  \begin{tabular}{llccc}
    \toprule
    Dataset & Metric & Top-1 & Top-10 & Top-20 \\
    \midrule
    NPLIB1 & Exact match$\uparrow$ & 13.94\% & 28.45\% & 30.58\% \\
    & Tanimoto$\uparrow$ & 0.5134 & 0.6404 & 0.6592 \\
    & Scaffold match$\uparrow$ & 38.95\% & 56.04\% & 57.51\% \\
    \midrule
    MassSpecGym & Exact match$\uparrow$ & 4.88\% & 14.53\% & 16.97\% \\
    & Tanimoto$\uparrow$ & 0.3567 & 0.4702 & 0.4954 \\
    & Scaffold match$\uparrow$ & 11.62\% & 24.66\% & 27.08\% \\
    \bottomrule
  \end{tabular}
\end{table*}

On NPLIB1, formula errors primarily reduce Top-1 ranking, while Top-20 exact match,
Tanimoto, and scaffold recovery remain 30.58\%, 0.6592, and 57.51\%. MassSpecGym is
more sensitive because the correct formula appears in MSBuddy's Top-5 for only 57.69\%
of spectra. Nevertheless, increasing the candidate budget raises exact match from
4.88\% to 16.97\% and scaffold recovery from 11.62\% to 27.08\%. Formula uncertainty
therefore affects prioritization most strongly, although chemically relevant hypotheses
often remain in the larger candidate set.

\section{Variation Across Random Seeds}
\label{app:seed_variation}
We conducted three independent runs of CoRe-Gen on both benchmarks.
Table~\ref{tab:three_seed_results} reports mean $\pm$ standard deviation,
showing consistent performance across random seeds.

\begin{table*}[t]
  \centering
  \caption{Performance of CoRe-Gen across three independent runs. The MassSpecGym row uses the released DiffMS encoder checkpoint.}
  \label{tab:three_seed_results}
  \small
  \resizebox{\textwidth}{!}{%
  \begin{tabular}{lcccccc}
    \toprule
    & \multicolumn{3}{c}{Top-1} & \multicolumn{3}{c}{Top-10} \\
    \cmidrule(lr){2-4}\cmidrule(lr){5-7}
    Model & Accuracy$\uparrow$ & MCES$\downarrow$ & Tanimoto$\uparrow$ & Accuracy$\uparrow$ & MCES$\downarrow$ & Tanimoto$\uparrow$ \\
    \midrule
    CoRe-Gen (NPLIB1) & $21.49\pm0.29$\% & $5.57\pm0.12$ & $0.57\pm0.03$ & $35.42\pm0.40$\% & $4.03\pm0.19$ & $0.67\pm0.04$ \\
    CoRe-Gen$^\dagger$ (MassSpecGym) & $20.02\pm0.26$\% & $16.09\pm0.45$ & $0.40\pm0.02$ & $22.45\pm0.30$\% & $13.90\pm0.33$ & $0.47\pm0.03$ \\
    \bottomrule
  \end{tabular}%
  }
\end{table*}

\section{Differentiable Fingerprint-to-Memory Bridge}
\label{app:fp_memory_bridge}
\begin{figure*}[t]
  \centering
  \includegraphics[width=0.78\textwidth]{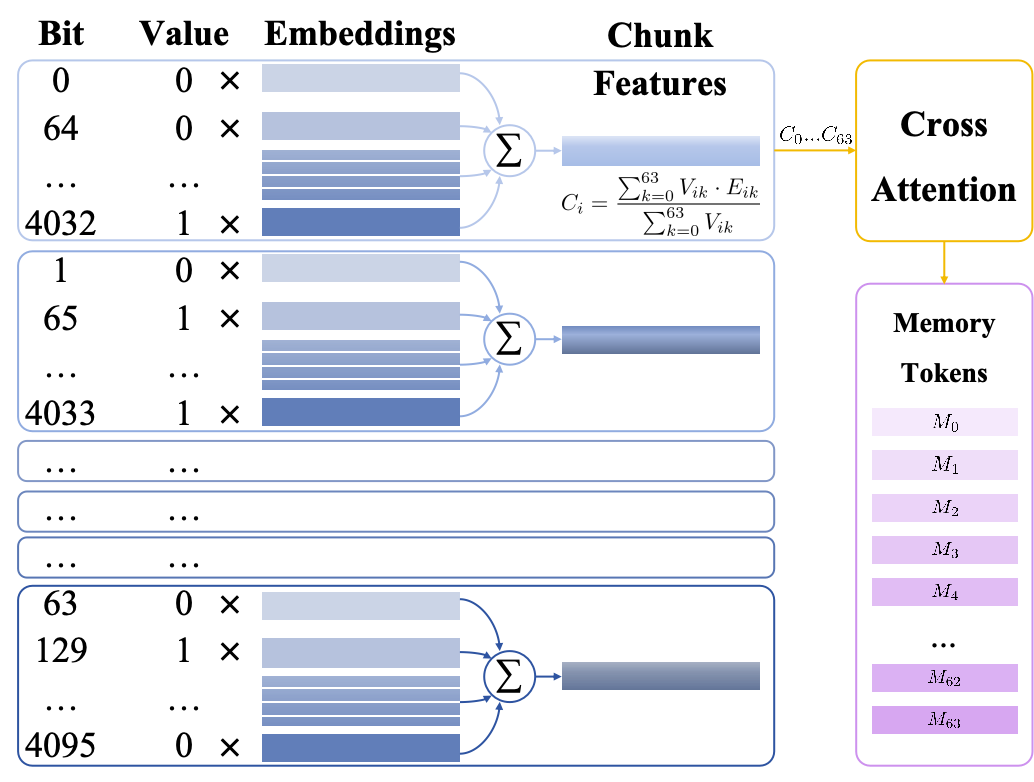}
  \caption{Differentiable fingerprint-to-memory bridge. The 4096 soft fingerprint values
  are assigned to 64 modulo chunks. Within each chunk, bit-specific embeddings are
  weighted by their predicted values and normalized to form a chunk feature. The
  contextualized chunk sequence provides 64 memory tokens for decoder cross-attention.}
  \label{fig:fp_memory_bridge}
\end{figure*}

Autoregressive fingerprint decoders commonly convert a binary fingerprint into a
variable-length sequence containing only active bit indices. For a predicted
fingerprint, this requires a hard threshold and index-selection operation. The resulting
sequence is suitable for a language-model encoder, but discards uncertainty in inactive
or borderline bits and blocks gradients from molecular token loss to the spectrum
encoder. CoRe-Gen replaces this handoff with the bridge in
Figure~\ref{fig:fp_memory_bridge}.

Let $\hat{\mathbf f}\in[0,1]^{4096}$ be the encoder output and let $C=64$. Modulo chunk
$i$ contains indices $i+kC$ for $k\in\{0,\ldots,63\}$. With a learned embedding
$\mathbf e_j\in\mathbb R^{512}$ for each fingerprint bit, its initial chunk feature is
\begin{equation}
\mathbf c_i=
\frac{\sum_{k=0}^{63}\hat f_{i+kC}\mathbf e_{i+kC}}
{\sum_{k=0}^{63}\hat f_{i+kC}+\epsilon},
\qquad i=0,\ldots,63.
\label{eq:fp_to_chunk}
\end{equation}
This modulo assignment follows fingerprint-folding practice: adjacent fingerprint bits
are distributed across different chunks, while bits aggregated in the same chunk are
well separated in the original index space, reducing local feature confusion. The
weighted average then produces a fixed-length source sequence. Six 512-dimensional,
8-head Transformer layers
with RoPE contextualize $\{\mathbf c_i\}_{i=0}^{63}$ into memory
$\mathbf M\in\mathbb R^{64\times512}$, which conditions every autoregressive decoding
layer through cross-attention.

During decoder pretraining, the same bridge accepts clean or corruption-augmented
fingerprints. During final spectrum-conditioned training, it consumes soft predictions
directly. We jointly optimize
$\mathcal L_{\mathrm{joint}}=\mathcal L_{\mathrm{AR}}+
\lambda_{\mathrm{fp}}\mathcal L_{\mathrm{fp}}+
\lambda_{\mathrm{sent}}\mathcal L_{\mathrm{sent}}$.
Because all bridge operations are differentiable, $\mathcal L_{\mathrm{AR}}$ updates the
spectrum encoder, bridge, and autoregressive decoder end to end; the fingerprint loss
simultaneously preserves the quality of the intermediate chemical condition.

\clearpage

\section{Modified SELFIES Interface}
\label{app:modified_selfies}
\begin{figure}[t]
  \centering
  \includegraphics[width=0.8\columnwidth]{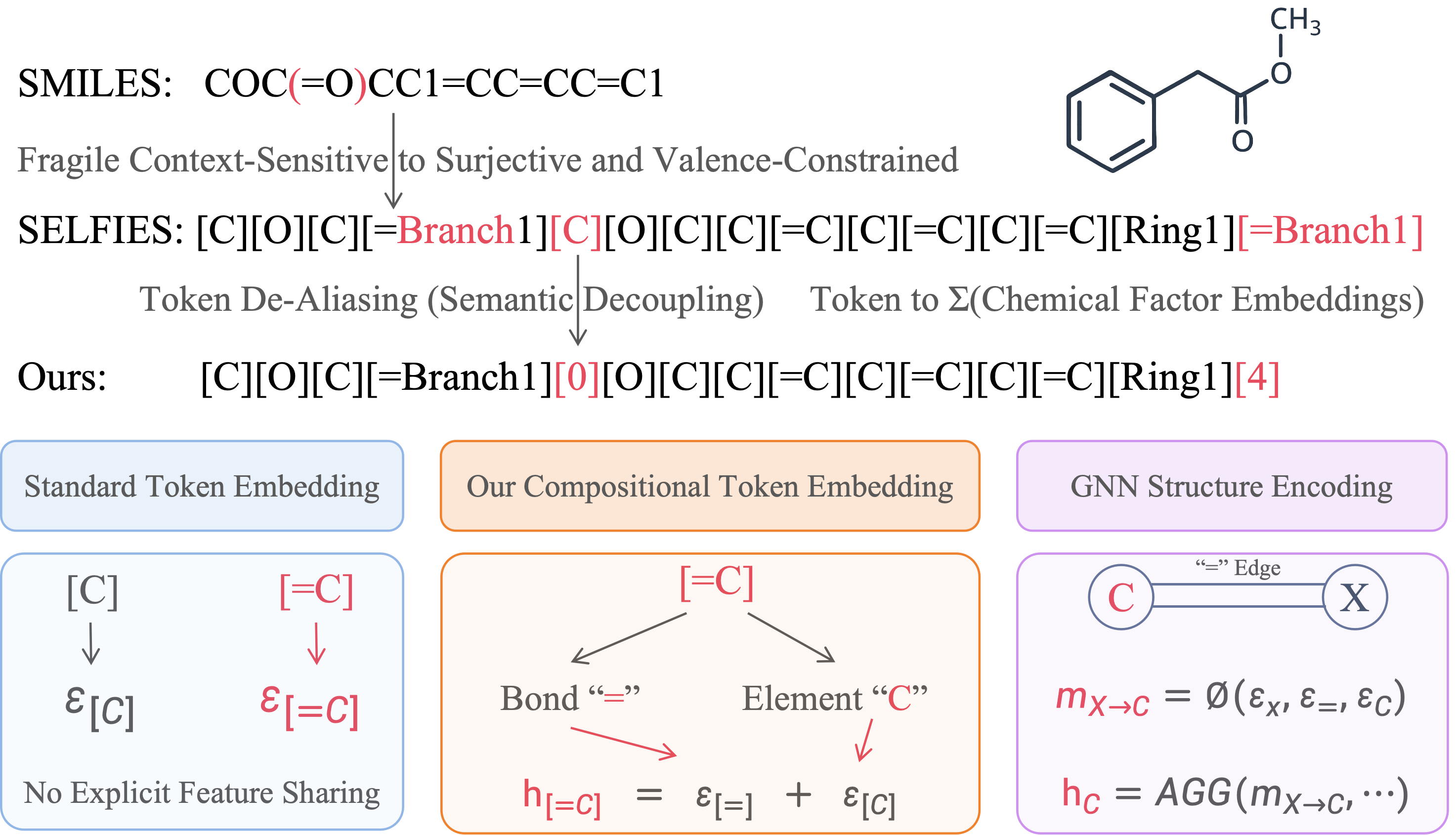}
  \caption{Illustration of the modified SELFIES token interface used for structure-aware decoding. We explicitly separate part type, element, bond prefix, ring/branch order, and hexadecimal continuation to reduce token-level ambiguity inherited from the original SELFIES logic.}
  \label{fig:structure_aware_decoding}
\end{figure}

\subsection{Token Factorization and Canonicalization Rules}
\label{app:token_rules}
Figure~\ref{fig:structure_aware_decoding} illustrates how we modify the original
SELFIES interface. Instead of treating each token as an atomic symbol with an unrelated
embedding vector, we decompose a token into several interpretable factors: its part type
(e.g., atom, ring, branch, or hexadecimal continuation), its chemical element when
applicable, its bond prefix, and its ring/branch order or hexadecimal index when
needed. In addition, we canonicalize the token representation before factorization so
that superficial variants do not create unnecessary vocabulary fragmentation. In
particular, we keep the core chemical identity needed for generation while avoiding a
proliferation of low-frequency token variants tied to explicit hydrogens or charges.
The interface also resolves a SELFIES-specific ambiguity: original SELFIES reuses
post-ring/branch tokens to encode hexadecimal continuation rather than atom semantics,
so the same surface token can act as either a structural-control symbol or a
chemical-identity symbol depending on context. We make this role explicit through the
part-type and digit factors.

This redesign serves two purposes. First, it reduces token-level ambiguity inherited
from standard SELFIES, where symbols that are chemically related can still look
independent to the decoder. By sharing factors across related tokens, the model can
reuse statistical strength across common local motifs such as bond type, atom identity,
and ring/branch behavior. Second, it improves robustness under limited supervision by
making the representation more compositional: a rare token can still benefit from common
factor embeddings even when its full token identity is infrequent in the finetuning
data.

An additional advantage of this factorized interface is that it exposes structured
subtargets on top of the autoregressive token prediction task. Because each target token
can be mapped to element-, bond-, ring-, and branch-level attributes, we can inject
lightweight structure supervision during decoder training without changing the decoding
architecture. This is the basis of the auxiliary sentence-level loss used in our decoder
pretraining and finetuning.

\subsection{Auxiliary Structure Supervision}
\label{app:embed_variants}
Given a target sequence $y_{1:T}$ and decoder logits over the modified SELFIES
vocabulary, we first compute the standard autoregressive cross-entropy loss. We then
derive additional factor-level supervision from the same logits by grouping vocabulary
items according to their structural attributes. Concretely, each target token induces an
element label, a bond label, a ring label, and a branch label whenever the
corresponding attribute is defined. Let $p_{t,v}$ denote the softmax probability of
vocabulary item $v$ at time step $t$, and let $\mathcal{V}_{a}$ denote the subset of
tokens associated with a structural attribute $a$. For each attribute, we compare the
expected count under the model distribution against the count implied by the target
sequence. Let $\mathcal{L}_{\mathrm{ce}}$ denote the standard token cross-entropy. The
auxiliary sentence-level losses are defined as follows.

\paragraph{Element loss.}
For each heavy element $e$, we compute its predicted count by summing token
probabilities over all vocabulary items whose atom identity is $e$, and its target count
by counting how many target tokens belong to that element. Both predicted and target
element-count vectors are then normalized by their total mass, and the loss is the mean
squared error between the normalized vectors:
\[
\mathcal{L}_{\mathrm{elem}}
=
\mathrm{MSE}\!\left(
\frac{\hat{\mathbf{c}}^{\,\mathrm{elem}}}{\|\hat{\mathbf{c}}^{\,\mathrm{elem}}\|_1+\varepsilon},
\frac{\mathbf{c}^{\mathrm{elem}}}{\|\mathbf{c}^{\mathrm{elem}}\|_1+\varepsilon}
\right).
\]
This term encourages the decoder to preserve the overall elemental composition pattern
even when the exact token sequence is uncertain.

\paragraph{Bond loss.}
We separately track the expected and target counts of double-bond and triple-bond
prefixed tokens. The bond loss is the sum of two mean-squared errors:
\[
\mathcal{L}_{\mathrm{bond}}
=
\mathrm{MSE}(\hat{c}_{=}, c_{=})
+
\mathrm{MSE}(\hat{c}_{\#}, c_{\#}).
\]
This term encourages the decoder to match the amount of higher-order bonding indicated
by the target sequence.

\paragraph{Ring and branch losses.}
Similarly, we define ring and branch losses by comparing the expected and target counts
of ring tokens and branch tokens:
\[
\mathcal{L}_{\mathrm{ring}}=\mathrm{MSE}(\hat{c}_{\mathrm{ring}}, c_{\mathrm{ring}}),
\qquad
\mathcal{L}_{\mathrm{branch}}=\mathrm{MSE}(\hat{c}_{\mathrm{branch}}, c_{\mathrm{branch}}).
\]
These terms encourage consistency in the predicted amount of ring closure and branching
behavior.

In our training setup, the active sentence-level rule set is
\texttt{element,bond,ring,branch}. Therefore, the auxiliary term used in optimization is
\[
\mathcal{L}_{\mathrm{sent}}
=
w_{\mathrm{elem}}\mathcal{L}_{\mathrm{elem}}
+ w_{\mathrm{bond}}\mathcal{L}_{\mathrm{bond}}
+ w_{\mathrm{ring}}\mathcal{L}_{\mathrm{ring}}
+ w_{\mathrm{branch}}\mathcal{L}_{\mathrm{branch}}.
\]
Our implementation also supports additional inequality and valence penalties, but these
terms are not enabled in the experiments reported in this paper.

The final training objective used in decoder pretraining is
\[
\mathcal{L}
=
\mathcal{L}_{\mathrm{ce}}
+ \lambda_{\mathrm{sent}} \mathcal{L}_{\mathrm{sent}},
\]
where $\lambda_{\mathrm{sent}}=0.1$ in our experiments. We enable this auxiliary term in
both decoder pretraining and decoder finetuning. The active factor set is
\texttt{element,bond,ring,branch}, with weights
$w_{\mathrm{elem}}=1.0$ and
$w_{\mathrm{bond}}=w_{\mathrm{ring}}=w_{\mathrm{branch}}=0.2$.

Operationally, this loss is computed from the same decoder logits as the main
autoregressive objective and therefore adds no extra decoding-time cost. Its role is not
to replace token prediction, but to bias the model toward recovering chemically
meaningful substructure attributes even when the exact full token is difficult to
predict. This is especially useful under noisy fingerprint conditioning, where the exact
token identity may be unstable but its structural factors remain partially recoverable.

\section{Evidence for the Predicted-Fingerprint Corruption Gap}
\label{app:corruption_gap}
\subsection{Clean-versus-predicted Fingerprint Gap}
\label{app:clean_vs_pred}
We define the predicted-fingerprint corruption gap as the mismatch between the clean
fingerprint extracted from the ground-truth molecule and the fingerprint predicted from
the input spectrum by the encoder. This gap matters because the decoder is pretrained on
fingerprint--molecule pairs, but at deployment time it receives predicted fingerprints
rather than oracle ones. If predicted fingerprints differ substantially from the clean
distribution, a decoder trained only on clean inputs will face a genuine train--test
condition mismatch. The test-set statistics below are a post-hoc diagnostic only: no
downstream test-set statistic was used to design the method, choose corruption
hyperparameters, or select a model. The motivating observation---that low-frequency
bits have higher prediction error---was established from training data before the
proposed components were developed.

We quantify this mismatch directly by comparing the number of active bits and the number
of bit errors between predicted and ground-truth fingerprints. For NPLIB1 test samples
($n=819$), predicted fingerprints are slightly denser than ground truth on average
($50.67$ active bits vs.\ $46.27$), with mean false-positive and false-negative counts
of $16.66$ and $12.26$, respectively. The mean total number of wrong bits is therefore
$28.92$. For our budget-preserving swap-style corruption process, the most relevant
quantity is $\mathrm{swap}\_k=\min(\mathrm{FP}, \mathrm{FN})$, which measures how many
bit drops and additions can be paired. On NPLIB1, this quantity has mean $10.97$,
median $9$, 90th percentile $23$, 95th percentile $27$, and is nonzero for $91.45\%$ of
test samples.

The mismatch is even stronger on MassSpecGym. After aligning by unique spectrum name, we
obtain 17,556 unique test samples. Here predicted fingerprints are slightly sparser than
ground truth on average ($57.78$ active bits vs.\ $61.31$), but the corruption itself is
substantially heavier: mean false-positive and false-negative counts are $29.43$ and
$32.96$, giving a mean total error count of $62.39$. The corresponding
$\mathrm{swap}\_k$ has mean $24.86$, median $24$, 90th percentile $37$, 95th percentile
$42$, and is nonzero for $99.97\%$ of test samples.

\begin{table}[t]
  \centering
  \caption{Clean-versus-predicted fingerprint mismatch on the downstream test sets.
  $\mathrm{swap}\_k=\min(\mathrm{FP},\mathrm{FN})$ summarizes the effective corruption
  budget relevant to our swap-style decoder corruption process.}
  \label{tab:clean_pred_gap}
  \scriptsize
  \setlength{\tabcolsep}{4pt}
  \resizebox{\columnwidth}{!}{%
  \begin{tabular}{lcccccccc}
    \toprule
    Dataset & $n$ & Pred ones & GT ones & FP & FN & Wrong & swap\_k mean & swap\_k p90 / p95 \\
    \midrule
    NPLIB1 & 819 & 50.67 & 46.27 & 16.66 & 12.26 & 28.92 & 10.97 & 23 / 27 \\
    MassSpecGym & 17,556 & 57.78 & 61.31 & 29.43 & 32.96 & 62.39 & 24.86 & 37 / 42 \\
    \bottomrule
  \end{tabular}
  }
\end{table}

These statistics support two design choices in CoRe-Gen. First, because many bits are
rare and receive limited direct supervision, improving fingerprint prediction itself is
important; this motivates encoder pretraining on large-scale synthetic spectra. Second,
even a strong encoder still produces fingerprints with substantial structured FP/FN
errors at test time, especially on MassSpecGym. This motivates corruption-aware decoder
pretraining, where the decoder is explicitly exposed to noisy fingerprint conditions
rather than only clean ones.

These downstream statistics were computed only after training to illustrate the
resulting condition mismatch. They did not determine the clipped-Poisson budget, its
clipping bounds, the corruption probability, or the training schedule. The final
hyperparameters are reported independently in Appendix~\ref{app:training_config}.

\subsection{Bit-level Error Statistics}
\label{app:bit_stats}
To characterize the fingerprint corruption pattern more precisely, we analyze errors at
the level of individual fingerprint bits and stratify them by training-set frequency.
The train-defined bit frequencies are computed on the downstream training split by
counting, for each Morgan bit, in how many training spectra that bit is present. In our
implementation, this frequency vector is also used to construct bit weights via
$w_b \propto 1/\sqrt{f_b+\epsilon}$ with $\epsilon=1.0$, followed by mean normalization
and clipping to $[0.1, 10.0]$. These weights are used for calibration/reweighting, while
the frequency statistics themselves provide a natural way to analyze which bits are easy
or hard to predict.

For analysis, we convert each bit count into an empirical training frequency and divide
bits into four train-defined buckets according to that frequency:
\textit{very rare} $[0,0.01)$, \textit{rare} $[0.01,0.05)$, \textit{mid} $[0.05,0.20)$,
and \textit{frequent} $[0.20,1.0]$. Intuitively, these buckets separate bits that are
seen in almost none of the training spectra from those that are observed regularly.
Since the learning signal for a bit is proportional to how often it appears in training,
this analysis directly tests whether sparsely observed substructures are indeed harder
for the encoder to recover.

\begin{table}[t]
  \centering
  \caption{MassSpecGym example of train-defined bit-frequency bucket analysis on the
  validation split. Lower-frequency bits are substantially harder to predict than
  frequently observed bits.}
  \label{tab:bit_bucket_msg}
  \scriptsize
  \setlength{\tabcolsep}{4pt}
  \resizebox{\columnwidth}{!}{%
  \begin{tabular}{lcccccc}
    \toprule
    Bucket & \# bits & Freq. range & Precision & Recall & F1 & TP / FP / FN \\
    \midrule
    Very rare & 3422 & $[0,0.01)$ & 0.198 & 0.083 & 0.117 & 69,350 / 280,274 / 767,242 \\
    Rare & 527 & $[0.01,0.05)$ & 0.314 & 0.285 & 0.299 & 184,132 / 402,700 / 462,108 \\
    Mid & 122 & $[0.05,0.20)$ & 0.428 & 0.581 & 0.493 & 279,174 / 373,700 / 200,920 \\
    Frequent & 25 & $[0.20,1.0]$ & 0.692 & 0.932 & 0.794 & 477,280 / 212,432 / 34,644 \\
    \bottomrule
  \end{tabular}
  }
\end{table}

Table~\ref{tab:bit_bucket_msg} shows a clear monotonic trend on MassSpecGym: the less
often a bit appears in the training set, the harder it is to predict reliably. The
effect is especially strong for very rare bits, which account for most of the 4096-bit
space (3422 bits) but achieve only 0.198 precision, 0.083 recall, and 0.117 F1. By
contrast, the frequent bucket contains only 25 bits but reaches 0.692 precision, 0.932
recall, and 0.794 F1. The false-negative count also drops sharply from 767,242 in the
very-rare bucket to only 34,644 in the frequent bucket.

This pattern supports our central motivation. In a finite training set, rare bits are
observed only a small number of times, so the encoder receives much weaker supervision
for the corresponding substructures. As a result, predicted fingerprints are not
corrupted uniformly: the corruption is concentrated on low-frequency, poorly trained
bits. This observation motivates encoder pretraining, whose role is to improve
prediction quality for weakly supervised bits by exposing the model to much broader
spectrum--substructure coverage. It also explains why decoder-side corruption should not
be modeled as uniform random noise: the dominant errors arise from structured,
frequency-dependent failures of the upstream predictor.

\section{Reliability, Coverage, and Practical Utility}
\label{app:reliability}
The following analyses characterize the reliability, coverage, and practical utility of
CoRe-Gen.

\subsection{Reliability versus Training-Set Coverage}
CoRe-Gen outputs should be treated as ranked structural hypotheses rather than
definitive identifications unless supported by reference spectra or orthogonal evidence
such as retention time, ion mobility, or NMR. To quantify where predictions are more
reliable, we group NPLIB1 test spectra by the maximum ECFP4 similarity of the reference
molecule to the training set.

\begin{table}[t]
  \centering
  \caption{NPLIB1 Top-1 exact match stratified by maximum structural similarity to the training set.}
  \label{tab:coverage_reliability}
  \small
  \begin{tabular}{lc}
    \toprule
    Maximum similarity to training set & Top-1 exact match$\uparrow$ \\
    \midrule
    $<0.4$ & 2.3\% \\
    $0.4$--$0.6$ & 11.2\% \\
    $0.6$--$0.8$ & 21.9\% \\
    $0.8$--$0.9$ & 35.3\% \\
    $>0.9$ & 36.4\% \\
    \bottomrule
  \end{tabular}
\end{table}

Accuracy is substantially higher in well-covered chemical regions; predictions far
from the training chemistry require correspondingly greater caution.

\subsection{Exact-Structure, Scaffold, and Family Recovery}
\label{app:scaffold_recovery}
We compare exact-structure recovery with canonical Bemis--Murcko scaffold recovery for
CoRe-Gen on NPLIB1. The correct scaffold appears for 44.20\%
of spectra at Top-1 and 57.43\% at Top-10, substantially more often than the exact
molecule at the corresponding candidate budgets.

\begin{table}[t]
  \centering
  \caption{Exact and Murcko-scaffold recovery of CoRe-Gen on NPLIB1.}
  \label{tab:scaffold_topk}
  \small
  \begin{tabular}{lcc}
    \toprule
    Top-$K$ & Exact match & Scaffold match \\
    \midrule
    Top-1 & 21.49\% & 44.20\% \\
    Top-10 & 35.42\% & 57.43\% \\
    \bottomrule
  \end{tabular}
\end{table}

We also identify several natural-product families in the NPLIB1 test set using CoRe-Gen
(Table~\ref{tab:family_recovery}); no
tropane alkaloids meet our filtering criteria.
Scaffold recovery is particularly strong for flavonoids and flavonoid glycosides. The
smaller subgroups should be interpreted cautiously because they contain few examples.

\begin{table}[t]
  \centering
  \caption{Top-20 recovery of CoRe-Gen for selected NPLIB1 compound families.}
  \label{tab:family_recovery}
  \scriptsize
  \setlength{\tabcolsep}{3.5pt}
  \begin{tabular}{lccc}
    \toprule
    Compound family & Number & Exact@20 & Scaffold@20 \\
    \midrule
    Flavonoid & 38 & 60.53\% & 76.32\% \\
    Flavonoid glycoside & 28 & 53.57\% & 71.43\% \\
    Benzylisoquinoline alkaloid & 3 & 0.00\% & 66.67\% \\
    Steroid/triterpenoid saponin & 5 & 0.00\% & 20.00\% \\
    \midrule
    Overall & 74 & 51.35\% & 70.27\% \\
    \bottomrule
  \end{tabular}
\end{table}

\subsection{Library-Depletion Experiment}
\label{app:library_depletion}
We simulate an out-of-library setting on 261 unique test molecules from five
NPClassifier superclass families. There is no exact molecule overlap between the
CoRe-Gen training and test sets, and 30.27\% of queries are scaffold-unseen relative to
training. For each query, we remove from the reference library every molecule with the
same InChIKey, the same canonical Bemis--Murcko scaffold, or ECFP4 Tanimoto similarity
$\geq0.6$. The library baseline uses the same benchmark-provided formula as CoRe-Gen,
exact-formula filtering, and modified-cosine ranking. CoRe-Gen uses its original
checkpoint, 100 candidates, and formula reranking without retraining or
experiment-specific tuning.

\begin{table}[t]
  \centering
  \caption{Out-of-library comparison after removing the true molecule, same-scaffold structures, and close analogues.}
  \label{tab:library_depletion}
  \small
  \begin{tabular}{lcc}
    \toprule
    Metric & Depleted library & CoRe-Gen \\
    \midrule
    Exact@1$\uparrow$ & 0 & 0.245 \\
    Exact@10$\uparrow$ & 0 & 0.398 \\
    Tanimoto@1$\uparrow$ & 0.312 & 0.616 \\
    Tanimoto@10$\uparrow$ & 0.338 & 0.721 \\
    MCES@1$\downarrow$ & 9.746 & 5.044 \\
    MCES@10$\downarrow$ & 8.765 & 3.398 \\
    Murcko@1$\uparrow$ & 0 & 0.464 \\
    Murcko@10$\uparrow$ & 0 & 0.598 \\
    No-hit rate$\downarrow$ & 0.502 & 0 \\
    \bottomrule
  \end{tabular}
\end{table}

Exact and Murcko recovery are zero by construction for the depleted library, so
Tanimoto, MCES, and no-hit rate are the informative comparisons. CoRe-Gen improves
Tanimoto by 0.304 at Top-1 and 0.383 at Top-10 and reduces MCES by 4.70 and 5.37,
respectively. Paired bootstrap confidence intervals over 10,000 resamples exclude zero
for all four comparisons. An oracle over all remaining library structures reaches only
0.338 Tanimoto and 8.762 MCES, nearly matching actual library-search Top-10 performance;
the principal limitation is therefore library coverage rather than spectral ranking.

\subsection{Candidate-Screening Workload Proxy}
\label{app:workload_proxy}
We use scaffold recovery and runtime as proxies for candidate-screening workload. For
100 randomly selected NPLIB1 test molecules, MetFrag~\citep{ruttkies2016metfrag} ranks PubChem isomers using each
spectrum and its benchmark-provided ground-truth formula, whereas CoRe-Gen generates its
Top-20 candidates.

\begin{table}[t]
  \centering
  \caption{Scaffold recovery and runtime on 100 randomly selected NPLIB1 queries.}
  \label{tab:workload_proxy}
  \small
  \begin{tabular}{lccccc}
    \toprule
    Method & Completed & Scaf.@1 & Scaf.@10 & Scaf.@20 & Runtime \\
    \midrule
    MetFrag & 95/100 & 27.37\% & 51.58\% & 62.11\% & $\sim$2 h 30 min \\
    CoRe-Gen & 100/100 & 42.00\% & 59.00\% & 62.00\% & $\sim$200 s \\
    \bottomrule
  \end{tabular}
\end{table}

MetFrag fails on five queries because of excessively large isomer sets. CoRe-Gen
completes all queries, is approximately $45\times$ faster, and achieves higher
Scaffold@1 and Scaffold@10, suggesting that correct structural cores are ranked earlier
and require less screening. A formal chemist user study is still required to measure
actual workload reduction.

\subsection{Post-Hoc Fingerprint-to-Token Attribution}
\label{app:interpretability}
We examine two links in the pipeline. MIST uses peak-to-substructure supervision, which
encourages peak evidence to enter chemically meaningful fingerprint features
\citep{goldman2023annotating}. For fingerprint-to-token attribution, at every decoding
step we ablate each active fingerprint bit and measure the change in token
log-probability. Influential shared ECFP4 bits are mapped to radius-2 atom environments,
and we test whether the generated atom lies in its most influential environment; control
tokens are excluded.

\begin{figure*}[t]
  \centering
  \includegraphics[width=0.82\textwidth]{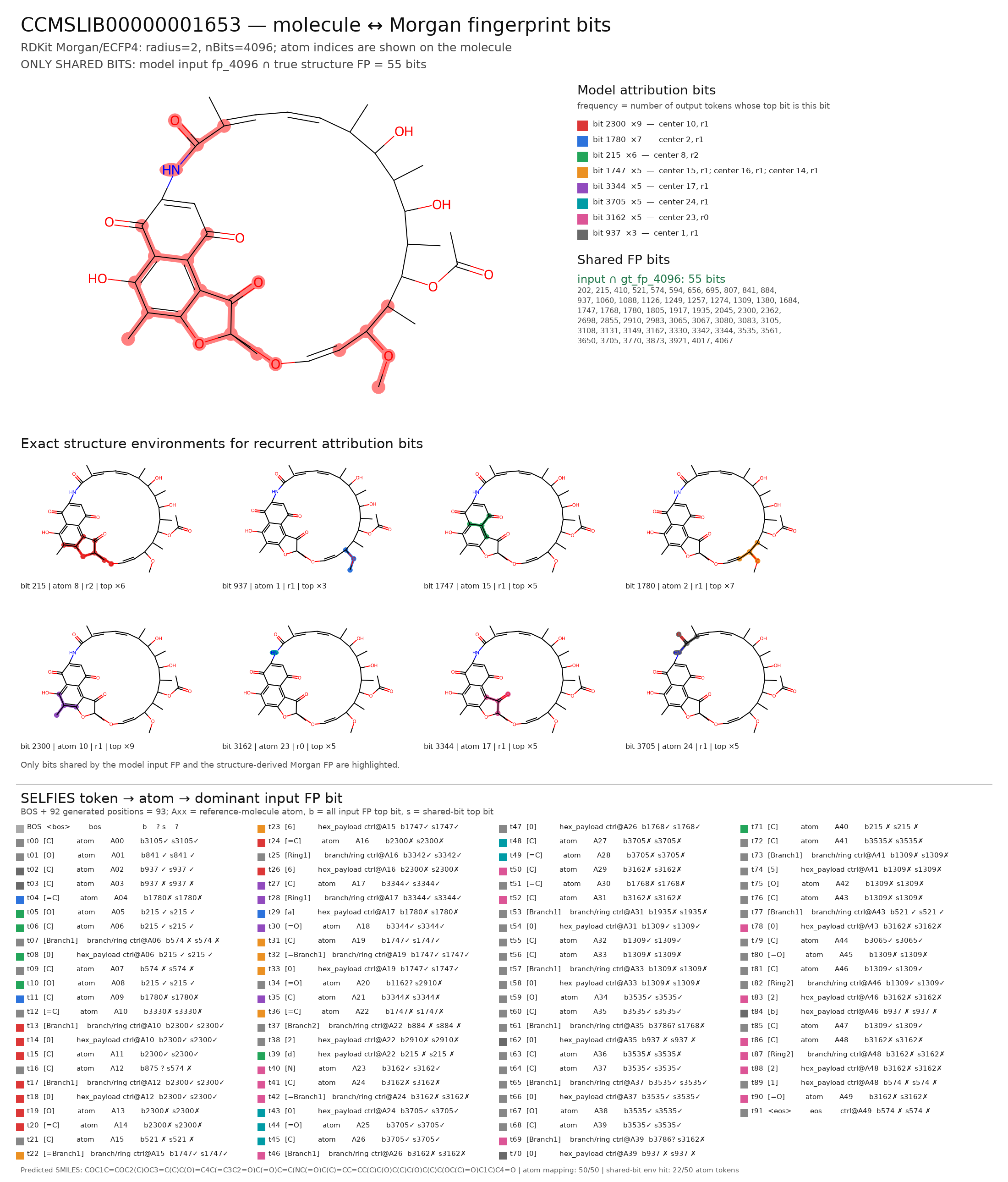}
  \caption{Fingerprint-to-token attribution case study. Recurrent influential shared
  ECFP4 bits are mapped to their radius-2 molecular environments, and each generated
  atom token is linked to its dominant input fingerprint bit.}
  \label{fig:fp_token_attribution}
\end{figure*}

In the random case study in Figure~\ref{fig:fp_token_attribution}, 22 of 50 atom tokens
(44.0\%) match the environment of their most influential bit, and the rate exceeds
60\% among the highest-contributing bits. This is a post-hoc example rather than a
comprehensive interpretability evaluation.

\subsection{Transfer beyond Simulator-Specific Artifacts}
\label{app:simulator_transfer}
Synthetic pretraining is intended to improve coverage of rare, long-tailed substructures
that receive little supervision from experimental datasets. To reduce dependence on any
single simulator, each molecule contributes five spectra: CFM-ID at 10, 20, and 40 eV,
ICEBERG, and SCARF. The encoder is then fine-tuned and evaluated only on experimental
spectra. Improvement on both downstream benchmarks therefore shows that at least part of
the learned signal transfers beyond the synthetic domain.

The current experiments do not isolate transferable chemical knowledge from
simulator-specific effects. Doing so requires controlled single-simulator, size-matched,
and leave-one-simulator-out ablations. We therefore treat simulator complementarity as a
limitation and leave those controlled comparisons to future work.

\section{Additional Failure Cases}
\label{app:failure_cases}
\begin{figure*}[t]
  \centering
  \includegraphics[width=0.98\textwidth]{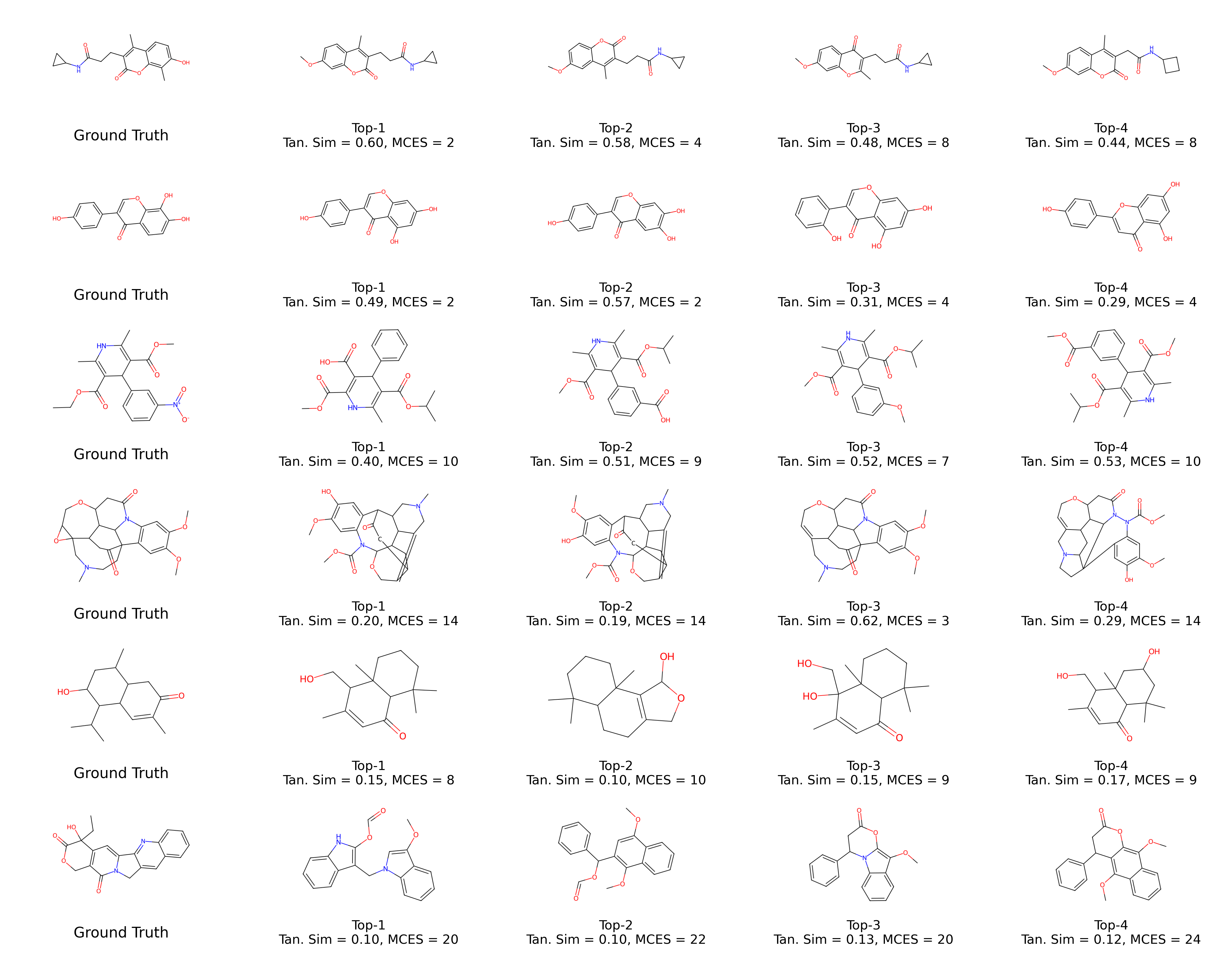}
  \caption{Additional representative failure cases. We show examples where generation
  errors arise from structurally similar analogues, isomeric ambiguity, or upstream
  fingerprint corruption.}
  \label{fig:qual_fail_appendix}
\end{figure*}

Figure~\ref{fig:qual_fail_appendix} presents additional representative failure cases of
CoRe-Gen.

\section{Limitations}
\label{app:limitations}
The current method pipeline still has several limitations:
\begin{itemize}
  \item \textbf{Encoder quality remains the main bottleneck.} The MIST-style spectrum encoder is still the dominant bottleneck of the pipeline. Our results show that synthetic data can partially alleviate the scarcity of real paired spectra, but the improvement is still not sufficient to fully close the predicted-fingerprint gap. Stronger spectrum encoders with better generalization, together with more high-quality real spectra, remain important directions.
  \item \textbf{Dependence on molecular formula information.} The main results use the benchmark-provided ground-truth formula for token masking and reranking. Appendix~\ref{app:formula_robustness} shows that MSBuddy-predicted formulas reduce performance, especially for Top-1 ranking and on MassSpecGym. This motivates both better formula predictors and generation methods less dependent on formula availability.
  \item \textbf{Hypotheses rather than definitive identifications.} Accuracy is lower in chemical regions distant from training data (Appendix~\ref{app:reliability}). Generated structures should therefore be validated with reference spectra or orthogonal evidence such as retention time, ion mobility, or NMR before being treated as identifications.
  \item \textbf{Speed-oriented single-pass generation.} CoRe-Gen is designed as a fast single-pass generation pipeline. In settings where accuracy is more important than latency, reference-based search, iterative recycling, or hybrid refinement strategies may be preferable, shifting the speed--accuracy trade-off toward higher precision at higher computational cost.
\end{itemize}

\section{Broader Impacts}
\label{app:broader_impacts}
This work advances de novo molecular structure elucidation from tandem mass spectrometry
(MS/MS), offering significant positive impacts for various scientific domains. By
improving the robustness of spectrum-to-structure generation under imperfect conditions,
our framework (CoRe-Gen) can substantially accelerate metabolomics profiling, natural
product discovery, and environmental monitoring, potentially aiding in the faster
development of life-saving therapeutics.

However, we must also acknowledge the dual-use potential of robust chemical generation
models. Like other computational tools for structural elucidation, our method could
theoretically be misused to reverse-engineer or identify illicit chemical compounds,
such as novel psychoactive substances (NPS) or harmful toxins, intended to evade
standard forensic detection. To mitigate these risks, we emphasize the importance of
responsible deployment and advocate for integrating such identification pipelines with
established chemical safety and security screening protocols.

\end{document}